\documentclass[letterpaper]{article} 
\usepackage{aaai25}  
\usepackage{times}  
\usepackage{helvet}  
\usepackage{courier}  
\usepackage[hyphens]{url}  
\usepackage{graphicx} 
\usepackage{natbib}  
\usepackage{caption} 
\usepackage{algorithm}
\usepackage{algorithmic}

\usepackage{amssymb}
\usepackage{amsmath}
\usepackage{amsthm}
\usepackage{graphicx}
 \usepackage{booktabs}

\newtheorem{proposition}{Proposition}[section]

\usepackage{newfloat}
\usepackage{listings}
\DeclareCaptionStyle{ruled}{labelfont=normalfont,labelsep=colon,strut=off} 
\floatstyle{ruled}
\newfloat{listing}{tb}{lst}{}
\floatname{listing}{Listing}
\title{Graph Structure Learning for Spatial-Temporal Imputation:\\Adapting to Node and Feature Scales}
\author{
    Xinyu Yang\textsuperscript{\rm 1}, Yu Sun\textsuperscript{\rm 1*}, Xinyang Chen\textsuperscript{\rm 2}\thanks{Corresponding authors.}, Ying Zhang\textsuperscript{\rm 1}, Xiaojie Yuan\textsuperscript{\rm 1}
}
\affiliations{
    \textsuperscript{\rm 1}College of Computer Science, DISSec, Nankai University, China\\
    \textsuperscript{\rm 2}School of Computer Science and Technology, Harbin Institute of Technology, Shenzhen,  China\\
    \{yangxinyu@dbis., sunyu@, yingzhang@, yuanxj@\}nankai.edu.cn,
     chenxinyang@hit.edu.cn \\
}

\usepackage{bibentry}

\begin{document}

\maketitle

\begin{abstract}
Spatial-temporal data collected across different geographic locations often suffer from missing values, posing challenges to data analysis. 
Existing methods primarily leverage fixed spatial graphs to impute missing values,
which implicitly assume that the spatial relationship is roughly the same for all features across different locations.
However, they may overlook the different spatial relationships of diverse features recorded by sensors in different locations.
To address this, we introduce the multi-scale \textbf{G}raph \textbf{S}tructure \textbf{L}earning framework for spatial-temporal \textbf{I}mputation (\textbf{GSLI}) that dynamically adapts to the heterogeneous spatial correlations.
Our framework encompasses node-scale graph structure learning to cater to the distinct global spatial correlations of different features, and feature-scale graph structure learning to unveil common spatial correlation across features within all stations.
Integrated with prominence modeling, our framework emphasizes nodes and features with greater significance in the imputation process. 
Furthermore, GSLI incorporates cross-feature and cross-temporal representation learning to capture spatial-temporal dependencies.
Evaluated on six real incomplete spatial-temporal datasets, GSLI showcases the improvement in data imputation.
\end{abstract}

%

\section{Introduction}
\begin{figure*}[t]
    \centering
    \includegraphics[width=\linewidth]{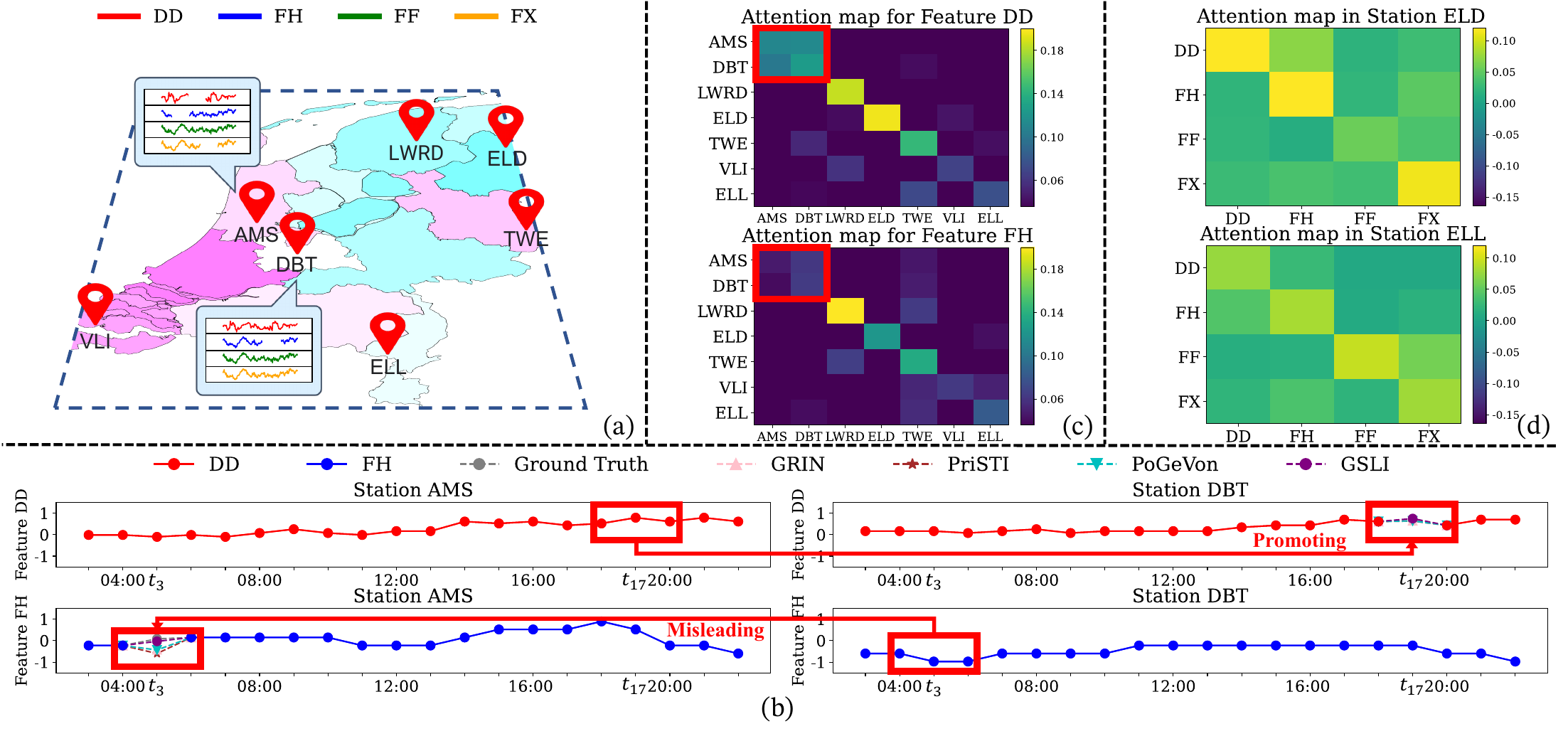}
    \vspace{-0.25in}
    \caption{(a) Incomplete spatial-temporal data with four features recorded in different stations in the Netherlands.
    (b) Imputation examples at timestamps $\mathit{t}_{3}$ and $\mathit{t}_{17}$.
    (c) The extracted attention maps for features DD and FH. 
    (d) The extracted attention maps for the four features in stations ELD and ELL.} 
    \label{fig:motivation}
\end{figure*}

The stations at different geographic locations may occur missing values when recording spatial-temporal data through multiple kinds of sensors \cite{TGCN,STDGAE}.
Figure \ref{fig:motivation}(a) presents an example of spatial-temporal data with four features recorded by seven stations located in the Netherlands.
As shown in Figure \ref{fig:motivation}(b), the readings of feature FH in station AMS at $t_3$ and feature DD in station DBT at $t_{17}$ are missing.
This may lead to anomalies in the patterns discovered by analysis models, thus creating challenges in mining spatial-temporal data \cite{DAMR}.

The spatial correlation of spatial-temporal data is usually given as a graph where stations are nodes and edges indicate geographic distance or connectivity between stations \cite{DCRNN,GWN,STDGAE}.
Existing works typically utilize the given graph and graph neural networks (GNN) to capture spatial dependencies  \cite{GRIN,PRISTI}.
Since AMS and DBT stations are geographically closest to each other, their corresponding edge weights in the graph are larger. 
Therefore, as shown in Figure \ref{fig:motivation}(b), we can get an accurate imputation result of the feature DD in DBT, according to the DD value in AMS station.
Unfortunately, this is not always the case, and we can observe that using the FH value in DBT station will mislead the existing methods for imputing the FH value in AMS station.
The reason is that features DD and FH are recorded by sensors from different domains, where DD captures wind direction data and FH is related to wind speed.
There is a significant correlation between wind direction information from AMS and DBT stations as they are located close to each other. 
However, since AMS is situated in an airport with a relatively sparse environment and DBT is located in a municipality with many buildings, the relationship between wind speed values of two stations is not clear.

Figure \ref{fig:motivation}(c) illustrates the spatial relationships of different features across all stations, where we extract the attention map from the cross-feature self-attention mechanism.
As shown, the spatial relationships for different features are varied, unlike the implicit assumption by existing methods that spatial relationships across stations are similar for different features.
In addition, in Figure \ref{fig:motivation}(d), we also find that there is a relatively fixed spatial relationship between the features within each station.
Specifically, the attention maps between features within the two stations ELD and ELL are generally similar.
This suggests that there also exist correlations between different features across all stations.
However, this correlation cannot be reflected in the given graph and thus cannot contribute to the imputation of existing methods.

Enlightened by the aforesaid analysis, we consider the multi-scale \textbf{G}raph \textbf{S}tructure \textbf{L}earning framework for spatial-temporal \textbf{I}mputation (\textbf{GSLI}).  Our main contributions can be summarized as follows.

(1) We present the node-scale graph structure learning to model the fine-grained global spatial correlations of different features. By adaptively learning independent global graph structures for different features, our framework can mitigate the negative effects between features in different domains and improve imputation performance.

(2) We design feature-scale graph structure learning to learn the common spatial correlation of different features over all nodes. Our framework can capture the spatial dependencies between features across each station with the help of the learned feature-scale graph structure.

(3) We incorporate prominence modeling into the graph structure learning processes to account for the varying influence of different nodes and different features. Thus, the nodes and features that contribute more to imputation can get stronger weights in the graph structures.

Experimental evaluations over real-world incomplete datasets demonstrate the superiority of our GSLI, by utilizing cross-feature representation learning and cross-temporal representation learning.

\section{Related Work}
\paragraph{Spatial-Temporal Imputation}
Spatial-temporal data, when viewed as multivariate time series by disregarding spatial correlation, often undergoes imputation using time series imputation methods.
Time series imputation utilizes various methods, including statistical approaches like mean imputation \cite{mean11}, last observation carried forward \cite{LastObsevation}, and local interpolation \cite{acuna2004treatment}, alongside techniques such as TRMF \cite{TRMF}, BTMF \cite{BTMF}, and TIDER \cite{TIDER} which employ low-rank matrix factorization.
RNN-based methods  GRU-D \cite{GRUD} and BRITS \cite{BRITS}, along with GAN-integrated methods GAN-2-Stage \cite{GAN18}, E$^2$GAN \cite{GAN19}, SSGAN \cite{SSGAN}, as well as self-attention and convolutional approaches in STCPA \cite{STCPA}, SAITS \cite{SAITS}, and TimesNet \cite{TimesNet} emphasize temporal dependencies.
Moreover, VAEs in MIWAE \cite{MIWAE}, GP-VAE \cite{gpvae}, TimeCIB \cite{TimeCIB}, diffusion models in CSDI \cite{CSDI}, MIDM \cite{MIDM}, SSSD \cite{SSSD}, and GPT4TS \cite{GPT4TS} using large language models are explored.
These methods, however, typically ignore the spatial adjacency crucial for spatial-temporal data, indicating potential improvements in the spatial dependency modeling.

For spatial-temporal imputation, LRTC-TNN \cite{LRTCTNN} uses low-rank tensor completion, GRIN  \cite{GRIN}  pioneers GNNs, and SPIN \cite{SPIN} targets error accumulation of GRIN for highly sparse data.
STD-GAE \cite{STDGAE} focuses on denoising graph autoencoders, while DAMR \cite{DAMR} dynamically extracts spatial correlations. PriSTI \cite{PRISTI} combines diffusion models with GNNs using GWN \cite{GWN}, learning rough graph structures.
PriSTI \cite{PRISTI} combines diffusion models with GNNs using GWN \cite{GWN}, learning rough graph structures.
Moreover, PoGeVon \cite{PoGeVon} predict missing values over both node time series
features and graph structures,
ImputeFormer \cite{ImputeFormer} and CASPER \cite{CASPER} utilize Transformer to capture spatial dependencies.
However, these methods often overlook the heterogeneity and common spatial dependencies among different features within nodes \cite{Chen23,Chen23-1,Chen24}.
On the contrary, our GSLI addresses feature heterogeneity via node-scale graph structure learning and prominence modeling, while also capturing spatial dependencies between features through feature-scale graph structure learning.

\paragraph{Spatial-Temporal Graph Structure Learning}
In the early stages of modeling spatial-temporal data, researchers commonly use the inherent graph structure and GNN to learn spatial dependencies\cite{DCRNN,TGCN}.
Pioneering the enhancement of spatial information within the given graph structure, GWN \cite{GWN} introduces graph structure learning by assigning two learnable embedding vectors to each node.
While methods like MTGNN \cite{MTGNN} and GTS \cite{GTS} design frameworks for learning discrete graph structures, 
AGCRN \cite{AGCRN} and CCRNN \cite{CCRNN} further these advancements by incorporating node-specific convolutions and learning independent graph structures for each convolution layer, respectively. SLCNN \cite{SLCNN} aimed to understand both global and local structural information in spatial-temporal data, whereas MegaCRN \cite{MegaCRN} adapted graph structures based on input signals.
CrossGNN \cite{crossgnn} utilizes graph structure learning to adapt to multiple scales of temporal periods, and heterogeneity between all variables in the forecasting tasks.
Unfortunately, these methods often focus on forecasting tasks, ignoring the crucial heterogeneity and correlation between features within nodes and differences in influence between nodes for the imputation task.
Compared to the forecasting task, imputing missing values is more difficult to capture temporal dependencies with incomplete observations, thus requiring learning accurate fine-grained spatial dependencies.
In contrast, our method captures feature-independent global spatial dependencies and spatial dependencies between features through the node and feature scales of graph structure learning, and reflects differences in node and feature influence by modeling prominence.


\section{Methodology}
In this section, we present the multi-scale Graph Structure Learning framework for spatial-temporal Imputation (GSLI).
The framework is built upon node-scale spatial learning, feature-scale spatial learning, cross-feature representation learning, and cross-temporal representation learning.

\begin{figure*}[t]
    \centering
    \includegraphics[width=\linewidth]{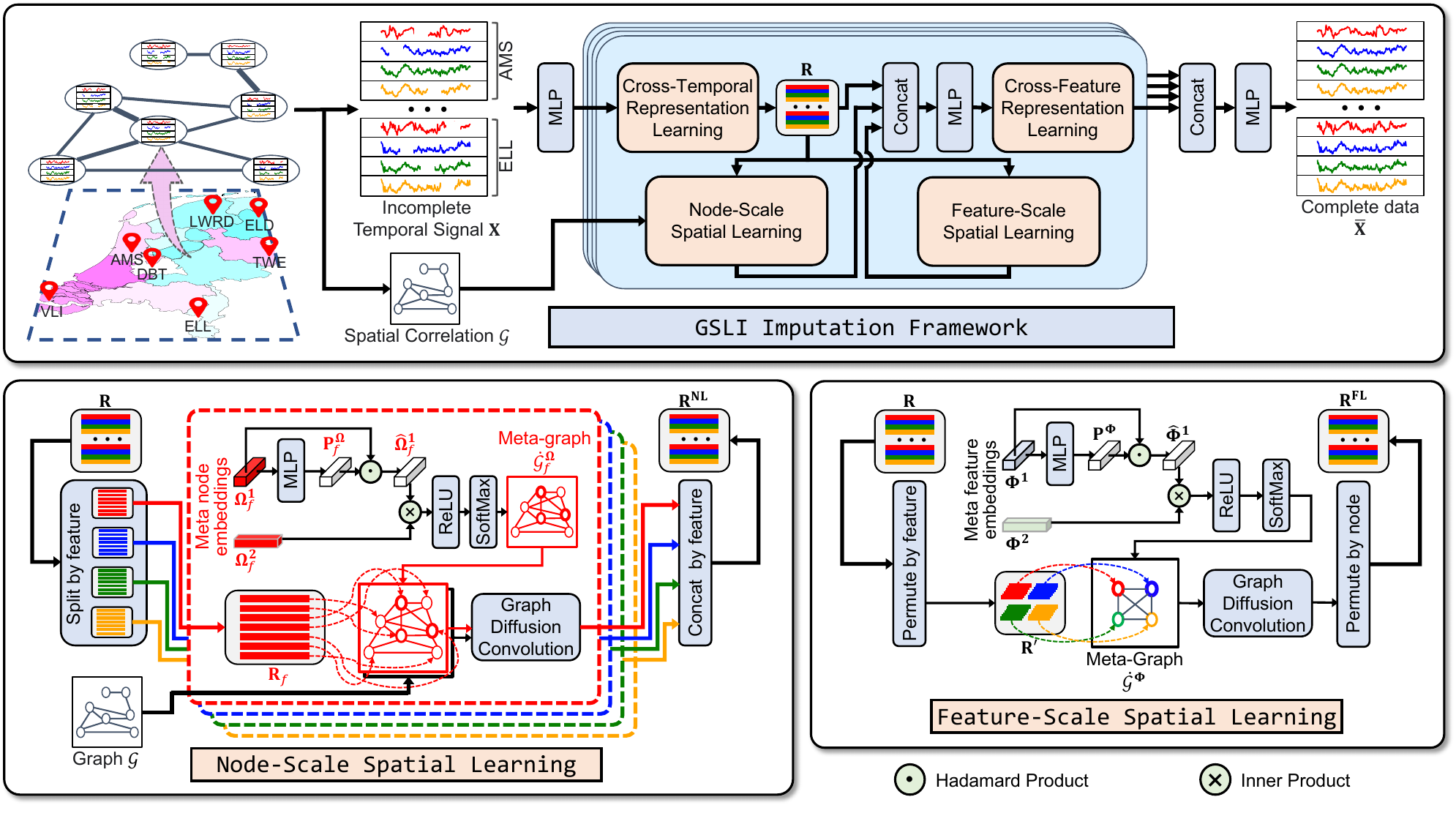}
    \vspace{-0.25in}
    \caption{The overview of multi-scale Graph Structure Learning framework for spatial-temporal Imputation (GSLI).
    GSLI incorporates node-scale spatial learning, which can adapt to feature heterogeneity, and feature-scale spatial learning, which can exploit correlations between features. 
    With cross-feature representation learning and cross-temporal representation learning, GSLI can effectively capture spatio-temporal dependencies for imputation.
    }
    \label{fig:framework}
\end{figure*}

\subsection{Problem Definition}
Spatial-temporal data $\left\langle \mathcal{G}, \mathbf{X}\right\rangle$ can be separated into two components: spatial correlation and temporal signal.
The spatial correlation is represented by a static graph $\mathcal{G}=\left(\mathcal{V},\mathcal{E}\right)$, where $\mathcal{V}$ is the set of $N$ nodes and $\mathcal{E}$ is the set of edges reflect the inherent relationships between nodes.
The adjacency matrix from $\mathcal{G}$ is denoted by $\mathbf{A} \in \mathbb{R}^{N\times N}$, where $A_{ij} \in \mathbf{A}$ reflects the weight of the edge $\left\langle v_i, v_j \right\rangle \in \mathcal{E}$, $v_i, v_j \in \mathcal{V}$.
The temporal signal $\mathbf{X} \in \mathbb{R}^{N \times T \times F}$ is the graph signal obtained by recording $T$ consecutive timestamps of $F$ features for each node over $\mathcal{G}$.
The missing status of $\mathbf{X}$ can be expressed by the mask matrix $\mathbf{M}\in \{0,1\}^{N \times T \times F}$.
If $m_{i,j,k} \in \mathbf{M}$ is equal to 0, it indicates that the observation $x_{i,j,k} \in \mathbf{X}$ is missing.
Given the incomplete spatial-temporal data, we aim to estimate all missing values $\mathbf{X} \odot \left( 1 - \mathbf{M}\right)$ in the temporal signal $\mathbf{X}$, where $\odot$ denotes the Hadamard product.

\subsection{Node-scale Spatial Learning}
To address the challenge of feature heterogeneity, i.e. the objects recorded by the sensors in the station are from different domains, we learn the node-scale graph structure for each feature independently to capture the global spatial dependencies between nodes, as shown in Figure \ref{fig:framework}.
We first split the input representation $\mathbf{R} \in \mathbb{R}^{N \times T \times F \times C}= \left\{ \mathbf{R}_{f} \in \mathbb{R}^{N \times T \times C }\right\}_{f=1}^{F}$ into $F$ parts based on features, where $C$ is the channel number in the deep space for each feature.
Then, we adopt the canonical approach \cite{GWN} to assign two learnable meta node embeddings $\mathbf{\Omega}^{\mathbf{1}}_{f}, \mathbf{\Omega}^{\mathbf{2}}_{f} \in \mathbb{R}^{N \times d}$ to each feature.
We denote $\mathbf{\Omega}^{\mathbf{1}}_{f}$ as the source node embedding and $\mathbf{\Omega}^{\mathbf{2}}_{f}$ as the target node embedding.
Since the average attention scores for each station obtained through the cross-feature self-attention mechanism are different, it inspires us that different nodes influence the overall imputation differently\footnote{Please see Prominence Modeling section in Appendix \cite{GSLI} for details.}.
To account for the varying influence of different nodes on feature $f$, we use the source embedding to learn the prominence vector $\mathbf{P}_f^{\mathbf{\Omega}} \in \mathbb{R}^{N \times d}$ for each node,
\begin{equation}
\label{eq:prominence-node}
    \mathbf{P}_f^{\mathbf{\Omega}} = {\rm MLP} \left( \mathbf{\Omega}^{\mathbf{1}}_{f} \right).
\end{equation}
To keep the resulting shape not changing, we then utilize the Hadamard product to obtain the refined source embedding $\widehat{\mathbf{\Omega}}^{\mathbf{1}}_{f}$ with $\mathbf{P}_f^{\mathbf{\Omega}}$,
\begin{equation}
\label{eq:refine-node}
    \widehat{\mathbf{\Omega}}^{\mathbf{1}}_{f} = \mathbf{\Omega}^{\mathbf{1}}_{f} \odot \mathbf{P}_f^{\mathbf{\Omega}}.
\end{equation}
This means that edges sourced from highly influential nodes will carry stronger weight in the learned graph structure.
The meta-graph $\dot{\mathcal{G}}_{f}^{\mathbf{\Omega}}$ which represents the global spatial correlations specific to the feature $f$ can be obtained by:
\begin{equation}
\label{eq:node-scale-graph-learning}
    \dot{\mathbf{A}}_{f}^{\mathbf{\Omega}}= {\rm SoftMax} \left[ {\rm ReLU} \left( \widehat{\mathbf{\Omega}}^{\mathbf{1}}_{f} {\mathbf{\Omega}^{\mathbf{2}}_{f}}^{\top}\right)\right],
\end{equation}
where $\dot{\mathbf{A}}_{f}^{\mathbf{\Omega}}$ is the adjacency matrix of $\dot{\mathcal{G}}_{f}^{\mathbf{\Omega}}$,
${\rm ReLU}(\cdot)$ is applied to eliminate the weakly correlated edges of meta-graphs,
the ${\rm SoftMax}(\cdot)$ function is used to normalize the adjacency matrices of meta-graphs.

By using the adjacency matrices of both the input graph $\mathcal{G}$ and the meta-graph $\dot{\mathcal{G}}_f^{\mathbf{\Omega}}$, we can use graph diffusion convolution \cite{DCRNN} to capture the node-scale spatial dependencies of signal $\mathbf{R}_f$:
\begin{align}
\label{eq:node-scale-graph-conv}
    \mathbf{R}^{\mathbf{NL}}_{f} = 
    \Sigma_{k=0}^{K}
     &  \left[
    \dot{\mathbf{A}}_{f}^{\mathbf{\Omega}} \mathbf{R}_{f} \mathbf{\Theta}_{k,f}^{\mathbf{\Omega 1}} + 
    {\left({\mathbf{D}^{\mathbf{O}}}^{-1} \mathbf{A}\right)}^k \mathbf{R}_{f} \mathbf{\Theta}_{k,f}^{\mathbf{\Omega 2}} +
    \notag \right.\\ 
    & \left.
   \quad {\left({\mathbf{D}^{\mathbf{I}}}^{-1} \mathbf{A}^{\top}\right)}^k \mathbf{R}_{f} \mathbf{\Theta}_{k,f}^{\mathbf{\Omega 3}}
    \right],
\end{align}
where $K$ is the step number of the graph diffusion process, 
$\mathbf{\Theta}_{k,f}^{\mathbf{\Omega 1}}$,
$\mathbf{\Theta}_{k,f}^{\mathbf{\Omega 2}}$,
$\mathbf{\Theta}_{k,f}^{\mathbf{\Omega 3}} \in \mathbb{R}^{C \times C}$ are graph convolution kernels, 
$\mathbf{D}^\mathbf{O}$ and $\mathbf{D}^\mathbf{I}$ are the out-degree and in-degree matrices of $\mathbf{A}$, respectively.
The output $\mathbf{R}^{\mathbf{NL}} \in \mathbb{R}^{N \times T \times F \times C}$ of node-scale spatial learning is obtained by concatenating the graph diffusion convolution output $\mathbf{R}^{\mathbf{NL}}_{f}$ from each feature,
\begin{equation}
\label{eq:concat-node-scale}
    \mathbf{R}^{\mathbf{NL}} = {\rm Concat} \left( 
    \mathbf{R}^{\mathbf{NL}}_{1} \|
    \mathbf{R}^{\mathbf{NL}}_{2} \dots \|
    \mathbf{R}^{\mathbf{NL}}_{F}
    \right).
\end{equation}
This allows us to capture the spatial dependence of each feature independently and avoid the patterns corresponding to different features in this module interfering with each other.
For the node-scale spatial learning, the time complexity is $\mathcal{O}(FN^2TC+FNTC^2+FN^2d+FNd^2)$, the space complexity is $\mathcal{O}(FNTC+FC^2+FN^2+FNd+Fd^2)$.
Please refer to the Complexity Analysis section in Appendix \cite{GSLI} for a detailed analysis.


If node $i$ of the graph with feature heterogeneity satisfying the correlation weight of feature $f_1$ from node $j$ to node $i$ is $x$, and the correlation weight of feature $f_2$ from node $j$ to node $i$ is $y \neq x$, we find the canonical graph convolution cannot address the feature heterogeneity as follows.
\setcounter{proposition}{0}
\begin{proposition}
\label{prop:gnn}
The result of $\dot{\mathbf{A}}^{\mathbf{\Omega}} \mathbf{R} $ in the first term of the canonical graph diffusion convolution of the channel $c$ for $f_2$ feature at timestamp $t$ for the node $i$ is
\begin{equation*}
    a^{\mathbf{\Omega}}_{i1} r_{1,{f_2},c}+\dots + x r_{j,{f_2},c} + \dots + a^{\mathbf{\Omega}}_{iN} r_{N,{f_2},c},
\end{equation*}
which is in conflict with the expected result $(a^{\mathbf{\Omega}}_{i1} r_{1,{f_2},c}+\dots + y r_{j,{f_2},c} + \dots + a^{\mathbf{\Omega}}_{iN} r_{N,{f_2},c})$,
where $\dot{\mathbf{A}}^{\mathbf{\Omega}} \in \mathbb{R}^{N \times N}$ is the learned global graph structure,
$a^{\mathbf{\Omega}}_{ij} \in \dot{\mathbf{A}}^{\mathbf{\Omega}}$,
$r_{1,{f_2},c}, r_{N,{f_2},c},  r_{j,{f_2},c}\in \mathbf{R}$.
\end{proposition}
Under the same premise, our node-scale spatial learning can also adapt to feature heterogeneity,
thus mitigating the misleading of heterogeneous features in neighboring nodes for imputation.
\begin{proposition}
\label{prop:gsli}
The result of $\dot{\mathbf{A}}^{\mathbf{\Omega}}_{f} \mathbf{R}_f $  in the first term of Equation \ref{eq:node-scale-graph-conv}  of the channel $c$ for $f_2$ feature at timestamp $t$ for the node $i$ is capable to get the expected value:
\begin{equation*}
     a^{\mathbf{\Omega}}_{f_2,i,1} r_{f_2, 1}+ \dots + y r_{f_2,j} + \dots + a^{\mathbf{\Omega}}_{f_2,i,N} r_{f_2, N}.
\end{equation*}
where $a^{\mathbf{\Omega}}_{f_2,i,j} \in \dot{\mathbf{A}}^{\mathbf{\Omega}}_{f_2}$,
$r_{{f_2},1}, r_{{f_2},N},  r_{{f_2},j}\in \mathbf{R}_f$.
\end{proposition}
The proofs are based on the information flow analysis, with details in Appendix.Proofs section \cite{GSLI}.

\subsection{Feature-scale Spatial Learning}
For modeling the spatial correlation of different features over all nodes, we first define two learnable meta feature embeddings $\mathbf{\Phi^{1}}, \mathbf{\Phi^{2}} \in \mathbb{R}^{F \times d}$,
where $\mathbf{\Phi^{1}}$ is the source feature embedding and $\mathbf{\Phi^{2}}$ is the target feature embedding.
To reflect the influence and heterogeneity of different features, we model the prominence vector $\mathbf{P}^{\mathbf{\Phi}} \in \mathbb{R}^{F \times d}$ for each feature,
\begin{equation}
\label{eq:prominence-feature}
    \mathbf{P}^{\mathbf{\Phi}} = {\rm MLP} \left( \mathbf{\Phi}^{\mathbf{1}} \right).
\end{equation}
Next, we refine the source feature embedding $\mathbf{\Phi^{1}}$ by:
\begin{equation}
\label{eq:refine-feature}
    \widehat{\mathbf{\Phi}}^{\mathbf{1}} = \mathbf{\Phi}^{\mathbf{1}} \odot \mathbf{P}^{\mathbf{\Phi}}.
\end{equation}
The source feature embeddings for features that are less heterogeneous from other features and can contribute to imputing missing values in other features will have stronger weights.
With the inner product between the embeddings, we can learn the meta-graph $\dot{\mathcal{G}}^{\mathbf{\Phi}}$ which represents the common spatial correlation of different features within each node,
\begin{equation}
    \dot{\mathbf{A}}^{\mathbf{\Phi}} = {\rm SoftMax} \left[ {\rm ReLU} \left( 
    \widehat{\mathbf{\Phi}}^{\mathbf{1}}  {\mathbf{\Phi}^{\mathbf{2}}}^{\top}
    \right)\right],
\end{equation}
where $\dot{\mathbf{A}}^{\mathbf{\Phi}}$ denotes the adjacency matrix of $\dot{\mathcal{G}}^{\mathbf{\Phi}}$.

To capture the spatial dependencies between different features,
we first permute the input $\mathbf{R} \in \mathbb{R}^{N \times T \times F \times C }$ into $\mathbf{R}^{\prime} \in \mathbb{R}^{F \times N \times T \times C }$ according to the features,
\begin{equation}
    \mathbf{R}^{\prime} = {\rm Permute}_{(2,0,1,3)} \left( \mathbf{R} \right).
\end{equation}
Then we obtain ${\mathbf{R}^{\prime}}^{\mathbf{FL}}$ through the graph diffusion convolution layer based on $\dot{\mathcal{G}}^{\mathbf{\Phi}}$:
\begin{equation}
    {\mathbf{R}^{\prime}}^{\mathbf{FL}} = \Sigma_{k=0}^{K}
    \dot{\mathbf{A}}^{\mathbf{\Phi}} \mathbf{R}^{\prime}
    \mathbf{\Theta}^{\mathbf{\Phi}}_{k},
\end{equation}
where $\mathbf{\Theta}^{\mathbf{\Phi}}_{k} \in \mathbb{R}^{C \times C}$ are graph convolution kernels.
The output of the feature-scale spatial learning $\mathbf{R}^{\mathbf{FL}} \in \mathbb{R}^{N \times T \times F \times C }$ can obtain from permuting the output of the diffusion convolution layer according to the nodes of the input data,
\begin{equation}
    \mathbf{R}^{\mathbf{FL}} =  {\rm Permute}_{(1,2,0,3)} \left({\mathbf{R}^{\prime}}^{\mathbf{FL}} \right).
\end{equation}

\subsection{Cross-Feature Representation Learning}
The goal of this phase is to self-adaptively obtain the representation that captures spatial dependencies between features across different nodes, which can be challenging to model a large number of spatial dependencies, i.e., $(N \times F)^2$.
To overcome this challenge, we use the input representation $\mathbf{R}$, with the outputs $\mathbf{R^{NL}}$ and $\mathbf{R^{FL}}$ obtained from the two scales of spatial learning, as inputs for this stage.

We start by concatenating the three inputs based on features and fusing them using MLP to obtain $\mathbf{E} \in \mathbb{R}^{N \times T \times F \times C}$,
\begin{equation}
    \mathbf{E} = {\rm MLP} \left[ 
    {\rm Concat} \left( \mathbf{R} \| \mathbf{R^{NL}} \| \mathbf{R^{FL}} \right)
    \right].
\end{equation}
Then we split $\mathbf{E} = \left\{ \mathbf{E}_{t} \in \mathbb{R}^{(N \times F) \times C }\right\}_{t=1}^{T}$ into $T$
segments according to timestamps and merge the node and feature dimensions of each segment.
Taking advantage of the Transformer \cite{transformer}, the learning process to obtain cross-feature representation of each timestamp $\mathbf{Z}_{t} \in \mathbb{R}^{(N \times F) \times C }$is:
\begin{equation}
    \mathbf{Z}_{t} = 
    {\rm SoftMax}\left( \frac{\mathbf{Q}_t {\mathbf{K}_t}^{\top} }{\sqrt{C}} \right)
    \mathbf{V}_t,
\end{equation}
where 
$\mathbf{Q}_t=\mathbf{E}_t \mathbf{W}_{Q}^{\mathbf{CF}}$,
$\mathbf{K}_t=\mathbf{E}_t \mathbf{W}_{K}^{\mathbf{CF}}$,
$\mathbf{V}_t=\mathbf{E}_t \mathbf{W}_{V}^{\mathbf{CF}}$,
$\mathbf{W}_{Q}^{\mathbf{CF}}, \mathbf{W}_{K}^{\mathbf{CF}}, \mathbf{W}_{V}^{\mathbf{CF}} \in \mathbb{R}^{C\times C}$
are learnable parameters.
Therefore, we can learn common spatial dependencies across different timestamps.
The output cross-feature representation $\mathbf{Z} \in \mathbb{R}^{N\times T \times F \times C}$ is the result of flattening the node and feature dimensions and concatenating each timestamp,
\begin{equation}
    \mathbf{Z} = {\rm Concat}
    \left[
    \left\{ 
    {\rm Flatten}_{(0,1)} \left(\mathbf{Z}_{t} \right)
    \right\}_{t=1}^{T}
    \right].
\end{equation}
The time complexity of this module is $\mathcal{O}(F^2NTC+FNTC^2+F^2d+Fd^2)$, the space complexity is $\mathcal{O}(FNTC+C^2+F^2+Fd+d^2)$.
For a detailed analysis, please see the Complexity Analysis section in Appendix \cite{GSLI}.

\subsection{Cross-Temporal Representation Learning}
Our goal in this stage is to capture temporal dependencies that can improve the performance of imputation.
Since capturing temporal dependencies on the original input signal is more reliable \cite{DBLP:conf/aaai/ZhangZQ17, LIM20211748}, we utilize the input temporal signal $\mathbf{X}^{\mathbf{I}}$ of the spatial-temporal data as the input.

We first project $\mathbf{X}^{\mathbf{I}}$ into deep space to obtain $\mathbf{H} \in \mathbb{R}^{N\times T \times F \times C}$,
\begin{equation}
    \mathbf{H} = {\rm MLP}\left( \mathbf{X}^{\mathbf{I}} \right).
\end{equation}
For capturing dependencies between different timestamps, we split $\mathbf{H} = \left\{ \mathbf{H}_{y} \in \mathbb{R}^{T \times C }\right\}_{y=1}^{(N \times K)}$ into $(N \times K)$ segments according all features across all nodes.
Next, we can obtain the cross-temporal representation of a feature within a node $\mathbf{R}_{y} \in \mathbb{R}^{T \times C }$ by
\begin{equation}
    \mathbf{R}_{y} = {\rm SoftMax}\left( \frac{\mathbf{Q}_y {\mathbf{K}_y}^{\top} }{\sqrt{C}} \right)
    \mathbf{V}_y,
\end{equation}
where
$\mathbf{Q}_y=\mathbf{H}_y \mathbf{W}_{Q}^{\mathbf{CT}}$,
$\mathbf{K}_y=\mathbf{H}_y \mathbf{W}_{K}^{\mathbf{CT}}$,
$\mathbf{V}_y=\mathbf{H}_y \mathbf{W}_{V}^{\mathbf{CT}}$.
To get the cross-feature representation $\mathbf{R}$, we need to concatenate all $\mathbf{R}_{y}$  and flatten them based on the node to which the features belong,
\begin{equation}
    \mathbf{R} = {\rm Flatten}_{(0,3)}
    \left[
    {\rm Concat} 
    \left(
    \left\{ 
    \mathbf{R}_{y} 
    \right\}_{y=1}^{N \times F}
    \right)
    \right].
\end{equation}
Therefore, we can learn common temporal dependencies across different features.

\subsection{The Framework of GSLI}
In this section, we introduce the multi-scale Graph Structure Learning framework for spatial-temporal data Imputation (GSLI).
The framework mainly consists of multiple layers with the same architecture. 
Each layer incorporates our proposed node-scale spatial learning, feature-scale spatial learning, cross-feature representation learning, and cross-temporal representation learning.
These components work together to capture the spatial-temporal dependencies required for accurate imputation.
Following previous studies \cite{CSDI,PRISTI,ImputeFormer}, we first learn temporal dependencies and then learn spatial dependencies .

When training the framework, it's impossible to know the ground truth of real missing values.
Therefore, we randomly selected some observations from $\mathbf{X}$ as the training label $\mathbf{X}^{\mathbf{B}} \in \mathbf{X}$, $\mathbf{X}^{\mathbf{B}} \in \mathbb{R}^{N\times T \times F \times C}$.
We use $\mathbf{M}^{\mathbf{B}}$ to represent the missing status of $\mathbf{X}^{\mathbf{B}}$
and compose the remaining observations as the input signal $\mathbf{X}^{\mathbf{I}} = \mathbf{X} \setminus \mathbf{X}^{\mathbf{B}}$ to the framework.
Then, we train GSLI by minimizing $\mathcal{L}$:
\begin{equation}
    \mathcal{L} = \mathbb{E} \left\| 
    \left( \mathbf{X}^{\mathbf{B}} - \overline{\mathbf{X}}\right) \odot \mathbf{M}^{\mathbf{B}}
    \right\|^2_2,
\end{equation}
where $ \overline{\mathbf{X}}$ is the output of the framework.
Note that to encourage the framework to focus on more diverse temporal and spatial dependencies and enhance adaptability and flexibility, we select different $\mathbf{X}^{\mathbf{B}}$ for each training step.

When imputing incomplete spatial-temporal data, we utilize the original temporal signal as the input signal $\mathbf{X}^{\mathbf{I}} = \mathbf{X}$.
The final estimated missing value we obtain is $\overline{\mathbf{X}} \odot \left( 1-\mathbf{M}\right)$.

\section{Experiment}
In this section, we evaluate the performance of our GSLI in imputation accuracy.
The experiments are conduct on a machine equipped with an Intel Xeon Silver 4314 2.40GHz CPU and an NVIDIA GeForce RTX 4090 24GB GPU.
The code and datasets are available online \shortcite{GSLI}\footnote{https://github.com/GSLI25/GSLI25/}.

\subsection{Experimental Setup}
\label{sec:exp-setup}

\paragraph{Datasets}
\begin{table}[t]
\centering
\renewcommand\tabcolsep{1pt}
     \renewcommand{\arraystretch}{0.7}
     \renewcommand\tabcolsep{1.5pt}
     \renewcommand{\arraystretch}{1}
     \resizebox{\linewidth}{!}{
        \begin{tabular}{@{}llllll@{}}
        \toprule
        Dataset & \#Nodes & \#Timestamps & \#Features & Missing & Type \\ \midrule
        DutchWind \cite{Dutch} & 7 & 8688 & 4 & 0.92\% & Wind \\
        BeijingMEO \cite{Beijing} & 18 & 8784 & 5 & 0.81\% & Meteo \\
        LondonAQ \cite{London} & 13 & 10897 & 3 & 13.81\% & Air Quality \\
        CN \cite{CN} & 140 & 2203 & 6 & 25.3\% & Air Quality \\
        Los \cite{Traffic} & 207 & 2016 & 1 & 1.25\% & Traffic \\
        LuohuTaxi \cite{Traffic} & 156 & 2976 & 1 & 24.76\% & Traffic \\ \bottomrule
        \end{tabular}
    }
    \vspace{-0.1in}
    \caption{Dataset summary}
    \label{tab:dataset-infor}
\end{table}

In our experiments, we use six spatial-temporal datasets that have real-world missing values.
Due to the unavailability of the ground truth of the missing values, we do not include them in the evaluation of imputation accuracy during comparative experiments, as noted in previous studies \cite{PRISTI,DAMR}. 
The main characteristics of these datasets are summarized in Table \ref{tab:dataset-infor}.
For spatial information, since DutchWind, BeijingMEO, and LondonAQ do not explicitly provide adjacency matrices,
we build adjacency matrices using the thresholded Gaussian kernel \cite{Gaussian} and the station coordinates following previous works \cite{PRISTI, DAMR}.

\paragraph{Baselines}
We compare with four state-of-the-art multivariate time series imputation methods: CSDI \cite{CSDI}, TimesNet \cite{TimesNet}, SAITS \cite{SAITS} and GPT4TS \cite{GPT4TS}, as well as seven spatial-temporal imputation methods: LRTC-TNN \cite{LRTCTNN}, GRIN \cite{GRIN}, STD-GAE \cite{STDGAE}, DAMR \cite{DAMR}, PriSTI \cite{PRISTI}, PoGeVon \cite{PoGeVon},
and ImputeFormer \cite{ImputeFormer}.

\subsection{Imputation Comparison}
We first conduct experiments with various missing rates and missing mechanisms to evaluate the imputation of GSLI.

\begin{table*}[t]
\centering
     \renewcommand\tabcolsep{1pt}
     \renewcommand{\arraystretch}{0.45}
     \resizebox{\linewidth}{!}{
     \renewcommand\tabcolsep{4pt}
\begin{tabular}{@{}lll|lllllllllllll@{}}
\toprule
Dataset & Missing rate & Metric & CSDI & TimesNet & SAITS & GPT4TS & LRTC-TNN & GRIN & STD-GAE & DAMR & PriSTI & PoGeVon & ImputeFormer & GSLI \\ \midrule
DutchWind & 10\% & RMSE & 0.464 & 0.510 & 0.473 & 0.560 & 0.586 & 0.437 & 0.473 & 0.609 & 0.483 & 0.493 & 0.526 & \textbf{0.410} \\
 &  & MAE & 0.223 & 0.318 & 0.265 & 0.366 & 0.293 & 0.229 & 0.251 & 0.398 & 0.217 & 0.287 & 0.299 & \textbf{0.205} \\  
 & 20\%  & RMSE & 0.490 & 0.620 & 0.482 & 0.669 & 0.611 & 0.441 & 0.490 & 0.611 & 0.489 & 0.472 & 0.542 & \textbf{0.421} \\
 &  & MAE & 0.241 & 0.428 & 0.272 & 0.468 & 0.316 & 0.234 & 0.266 & 0.404 & 0.227 & 0.256 & 0.313 & \textbf{0.213} \\ 
 & 30\%  & RMSE & 0.505 & 0.717 & 0.498 & 0.755 & 0.659 & 0.450 & 0.518 & 0.615 & 0.507 & 0.497 & 0.551 & \textbf{0.436} \\
 &  & MAE & 0.258 & 0.515 & 0.285 & 0.544 & 0.349 & 0.240 & 0.291 & 0.407 & 0.242 & 0.282 & 0.321 & \textbf{0.223} \\ 
 & 40\%  & RMSE & 0.526 & 0.787 & 0.504 & 0.817 & 0.705 & 0.455 & 0.564 & 0.621 & 0.537 & 0.569 & 0.565 & \textbf{0.448} \\
 &  & MAE & 0.284 & 0.578 & 0.293 & 0.600 & 0.392 & 0.247 & 0.337 & 0.411 & 0.269 & 0.364 & 0.330 & \textbf{0.234} \\   \midrule
BeijingMEO & 10\% & RMSE & 0.466 & 0.476 & 0.486 & 0.527 & 0.619 & 0.432 & 0.485 & 0.723 & 0.457 & 0.534 & 0.516 & \textbf{0.399} \\
 &  & MAE & 0.208 & 0.306 & 0.290 & 0.358 & 0.322 & 0.242 & 0.291 & 0.510 & 0.213 & 0.353 & 0.283 & \textbf{0.203} \\
 & 20\%  & RMSE & 0.478 & 0.528 & 0.494 & 0.634 & 0.658 & 0.438 & 0.494 & 0.730 & 0.472 & 0.541 & 0.527 & \textbf{0.407} \\
 &  & MAE & 0.217 & 0.365 & 0.303 & 0.483 & 0.343 & 0.248 & 0.297 & 0.517 & 0.229 & 0.347 & 0.295 & \textbf{0.210} \\
 & 30\%  & RMSE & 0.490 & 0.599 & 0.498 & 0.720 & 0.693 & 0.445 & 0.500 & 0.713 & 0.504 & 0.535 & 0.541 & \textbf{0.415} \\
 &  & MAE & 0.227 & 0.440 & 0.306 & 0.571 & 0.365 & 0.254 & 0.301 & 0.503 & 0.262 & 0.333 & 0.307 & \textbf{0.217} \\
 & 40\%  & RMSE & 0.506 & 0.674 & 0.506 & 0.788 & 0.734 & 0.453 & 0.508 & 0.716 & 0.562 & 0.562 & 0.541 & \textbf{0.423} \\
 &  & MAE & 0.240 & 0.514 & 0.316 & 0.636 & 0.390 & 0.262 & 0.306 & 0.509 & 0.301 & 0.364 & 0.311 & \textbf{0.224} \\  \midrule
LondonAQ & 10\% & RMSE & 0.298 & 0.406 & 0.375 & 0.481 & 0.490 & 0.311 & 0.597 & 0.721 & 0.314 & 0.375 & 0.402 & \textbf{0.272} \\
 &  & MAE & 0.182 & 0.264 & 0.249 & 0.321 & 0.310 & 0.198 & 0.471 & 0.493 & 0.192 & 0.232 & 0.262 & \textbf{0.173} \\
 & 20\%  & RMSE & 0.321 & 0.539 & 0.398 & 0.633 & 0.532 & 0.332 & 0.609 & 0.744 & 0.401 & 0.395 & 0.412 & \textbf{0.305} \\
 &  & MAE & 0.190 & 0.358 & 0.259 & 0.429 & 0.334 & 0.204 & 0.475 & 0.499 & 0.206 & 0.240 & 0.266 & \textbf{0.188} \\
 & 30\%  & RMSE & 0.340 & 0.658 & 0.413 & 0.741 & 0.584 & 0.351 & 0.665 & 0.786 & 0.502 & 0.415 & 0.436 & \textbf{0.320} \\
 &  & MAE & 0.199 & 0.449 & 0.265 & 0.508 & 0.367 & 0.213 & 0.510 & 0.522 & 0.234 & 0.250 & 0.275 & \textbf{0.191} \\
 & 40\%  & RMSE & 0.375 & 0.747 & 0.429 & 0.811 & 0.642 & 0.360 & 0.631 & 0.782 & 0.624 & 0.428 & 0.433 & \textbf{0.335} \\
 &  & MAE & 0.212 & 0.520 & 0.278 & 0.564 & 0.405 & 0.223 & 0.495 & 0.522 & 0.280 & 0.263 & 0.279 & \textbf{0.205} \\  \midrule
CN & 10\% & RMSE & 0.472 & 0.668 & 0.474 & 0.490 & 0.561 & 0.403 & 0.370 & 0.879 & 0.387 & 0.634 & 0.389 & \textbf{0.253} \\
 &  & MAE & 0.182 & 0.458 & 0.285 & 0.315 & 0.357 & 0.241 & 0.205 & 0.612 & 0.179 & 0.394 & 0.204 & \textbf{0.120} \\
 & 20\%  & RMSE & 0.436 & 0.698 & 0.481 & 0.509 & 0.596 & 0.417 & 0.388 & 0.884 & 0.423 & 0.633 & 0.401 & \textbf{0.267} \\
 &  & MAE & 0.194 & 0.485 & 0.289 & 0.333 & 0.387 & 0.251 & 0.216 & 0.622 & 0.197 & 0.394 & 0.210 & \textbf{0.129} \\
 & 30\%  & RMSE & 0.442 & 0.733 & 0.490 & 0.544 & 0.644 & 0.434 & 0.405 & 0.906 & 0.477 & 0.632 & 0.418 & \textbf{0.282} \\
 &  & MAE & 0.211 & 0.515 & 0.297 & 0.365 & 0.426 & 0.264 & 0.228 & 0.629 & 0.226 & 0.399 & 0.223 & \textbf{0.139} \\
 & 40\%  & RMSE & 0.465 & 0.766 & 0.499 & 0.593 & 0.706 & 0.451 & 0.425 & 0.905 & 0.512 & 0.582 & 0.444 & \textbf{0.299} \\
 &  & MAE & 0.233 & 0.545 & 0.304 & 0.409 & 0.471 & 0.277 & 0.243 & 0.629 & 0.262 & 0.349 & 0.237 & \textbf{0.150} \\  \midrule
Los & 10\% & RMSE & 0.311 & 0.531 & 0.535 & 0.397 & 0.501 & 0.295 & 0.945 & 0.513 & 0.293 & 0.365 & 0.445 & \textbf{0.263} \\
 &  & MAE & 0.177 & 0.339 & 0.292 & 0.257 & 0.332 & 0.188 & 0.541 & 0.363 & 0.186 & 0.209 & 0.227 & \textbf{0.159} \\
 & 20\%  & RMSE & 0.333 & 0.560 & 0.538 & 0.452 & 0.557 & 0.306 & 0.942 & 0.515 & 0.313 & 0.377 & 0.442 & \textbf{0.273} \\
 &  & MAE & 0.185 & 0.365 & 0.296 & 0.297 & 0.372 & 0.194 & 0.539 & 0.362 & 0.200 & 0.216 & 0.223 & \textbf{0.164} \\
 & 30\%  & RMSE & 0.355 & 0.602 & 0.544 & 0.543 & 0.618 & 0.319 & 0.944 & 0.521 & 0.353 & 0.390 & 0.459 & \textbf{0.282} \\
 &  & MAE & 0.193 & 0.401 & 0.299 & 0.356 & 0.415 & 0.201 & 0.541 & 0.368 & 0.225 & 0.223 & 0.230 & \textbf{0.168} \\
 & 40\%  & RMSE & 0.398 & 0.649 & 0.558 & 0.651 & 0.667 & 0.332 & 0.944 & 0.525 & 0.408 & 0.401 & 0.461 & \textbf{0.294} \\
 &  & MAE & 0.210 & 0.440 & 0.306 & 0.422 & 0.454 & 0.208 & 0.541 & 0.363 & 0.267 & 0.230 & 0.231 & \textbf{0.173} \\  \midrule
LuohuTaxi & 10\% & RMSE & 0.467 & 0.514 & 0.452 & 0.576 & 0.680 & 0.436 & 0.783 & 0.667 & 0.523 & 0.497 & 0.456 & \textbf{0.410} \\
 &  & MAE & 0.313 & 0.388 & 0.308 & 0.429 & 0.459 & 0.301 & 0.612 & 0.523 & 0.367 & 0.340 & 0.309 & \textbf{0.276} \\
 & 20\%  & RMSE & 0.468 & 0.540 & 0.455 & 0.692 & 0.718 & 0.439 & 0.780 & 0.671 & 0.525 & 0.498 & 0.453 & \textbf{0.414} \\
 &  & MAE & 0.315 & 0.415 & 0.312 & 0.503 & 0.489 & 0.304 & 0.612 & 0.527 & 0.363 & 0.342 & 0.308 & \textbf{0.279} \\
 & 30\%  & RMSE & 0.472 & 0.577 & 0.456 & 0.773 & 0.767 & 0.443 & 0.781 & 0.684 & 0.531 & 0.503 & 0.456 & \textbf{0.419} \\
 &  & MAE & 0.319 & 0.450 & 0.312 & 0.563 & 0.530 & 0.307 & 0.612 & 0.538 & 0.370 & 0.346 & 0.311 & \textbf{0.283} \\
 & 40\%  & RMSE & 0.474 & 0.621 & 0.461 & 0.828 & 0.780 & 0.448 & 0.781 & 0.684 & 0.537 & 0.511 & 0.456 & \textbf{0.424} \\
 &  & MAE & 0.323 & 0.488 & 0.317 & 0.608 & 0.548 & 0.311 & 0.613 & 0.539 & 0.374 & 0.351 & 0.311 & \textbf{0.287} \\
\bottomrule
\end{tabular}
    }
    \vspace{-0.1in}
    \caption{Imputation performance of GSLI compared to existing methods with various missing rates}
    \vspace{-0.1in}
    \label{tab:compare-exp}
\end{table*}

Since we can not access the ground truth of the missing values, we randomly remove different percentages of the observations as imputation labels for evaluation through the Missing Completely at Random (MCAR) mechanism \cite{MCAR}.
We evaluate the imputation performance using RMSE \cite{RMSE}
and MAE \cite{MAE}
, as suggested by previous studies \cite{STDGAE, PRISTI}.
For both metrics, a smaller value indicates a more accurate imputation.
We repeat each experiment five times and report the average results in Table \ref{tab:compare-exp}.
We can find that as the missing rate increases, the imputation performance of most methods decreases due to the reduction of information available to the imputation models
Additionally, we observe that the current spatial-temporal imputation methods do not outperform the multivariate time series imputation methods in a significant manner. This can be attributed to the imprecision of the given graph structure affects the modeling of spatial dependencies.
On the contrary, our GSLI achieves consistently superior performance to other methods over different datasets with various missing rates.
This is because our GSLI accurately models spatial dependencies through graph structure learning at both the node and feature scales.
Moreover, as we learn fine-grained spatial independent correlations for each feature, our method can outperform spatial-temporal imputation methods that only model inter-node dependencies.


Since the occurrence of missing data in real-world scenarios is usually related to the external environment or the sensors themselves, we additionally consider the missing at random (MAR) \cite{MAR} and missing not at random (MNAR) \cite{MNAR} mechanisms.
Following existing works \cite{GAIN,miao21marmnar}, we explore the imputation performance of different methods with various missing mechanisms.
Figure \ref{fig:missingmechanism_dutch} and
Missing Mechanisms section in Appendix \cite{GSLI}
illustrate the performance of different imputation methods with 10\% missing values.
We find that various imputation methods have similar performance levels with different missing mechanisms. 
As a result, we use MCAR by default in other experiments.
In addition, our GSLI method consistently delivers optimal results across different missing mechanisms. This indicates that the GSLI can adapt well to different real-life missing data scenarios.

\begin{figure}[t]
    \centering
    \includegraphics[width=\linewidth]{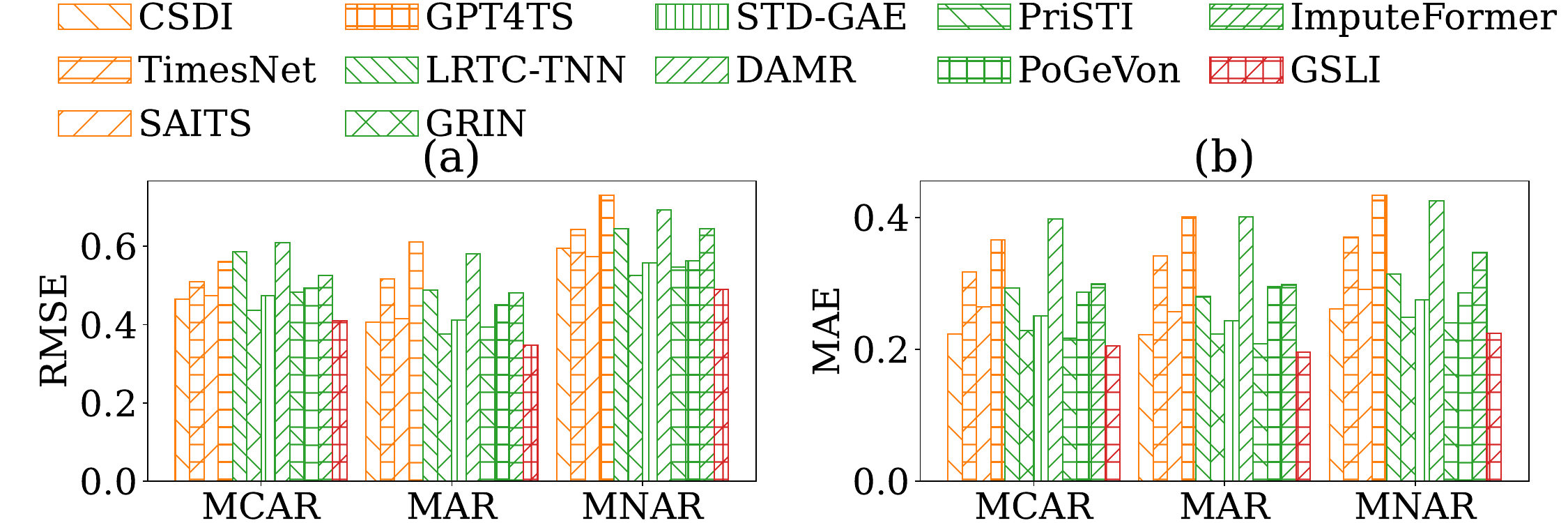}
    \vspace{-0.2in}
    \caption{Varying the missing mechanism over DutchWind dataset with 10\% missing values}
    \label{fig:missingmechanism_dutch}
\end{figure}

\subsection{Ablation Study}
\label{sec:ablation}
To validate the efficiency of each component of GSLI, we set up the following ablation variants:
\textbf{(1) TemporalGCN}: This variant utilizes cross-temporal representation learning to capture temporal dependencies, and the graph diffusion convolution to model spatial dependencies based on the existing adjacency matrix between nodes.
\textbf{(2) TemporalFeatrueRL}: The temporal dependencies are captured using cross-temporal representation learning, and spatial dependencies are captured using cross-feature representation learning.
\textbf{(3) w/o Cross-temporal}: We do not use cross-temporal representation learning to capture temporal dependencies in this situation.
\textbf{(4) w/o Cross-feature}: We do not utilize cross-feature representation learning to model spatial dependencies between features across different nodes.
\textbf{(5) w/o Feature-Split\&Scale}: We replace our node-scale spatial learning and feature-scale spatial learning with the canonical Graph Diffusion Convolution.
\textbf{(6) w/o Prominence}: When modeling spatial dependencies, the graph structure learning for both scales does not model the prominence of nodes in meta-graphs.
\textbf{(7) w/o Node-scale}: We only model the spatial correlation between different features for graph structure learning.
\textbf{(8) w/o Feature-scale}: We learn different node-scale graph structures for different features, and ignore the spatial correlation between features.
\textbf{(9) w/o GSL}: We do not perform any graph structure learning.
Instead, we use the graph diffusion convolution that takes the given adjacency matrix as input and the cross-feature representation learning to learn the spatial dependencies.

\begin{table}[t]

\centering
\renewcommand\tabcolsep{4pt}
     \renewcommand{\arraystretch}{1}
     \resizebox{\linewidth}{!}{
     \renewcommand\tabcolsep{2pt}
        \begin{tabular}{@{}l|ll|ll|ll|ll@{}}
        \toprule
        Method & \multicolumn{2}{l|}{DutchWind} & \multicolumn{2}{l|}{BeijingMEO} & \multicolumn{2}{l|}{LondonAQ} & \multicolumn{2}{l}{CN} \\ \cmidrule(l){2-9} 
         & RMSE & MAE & RMSE & MAE & RMSE & MAE & RMSE & MAE \\ \midrule
        TemporalGCN & 0.4453 & 0.2335 & 0.4175 & 0.2189 & 0.3133 & 0.2023 & 0.2989 & 0.1492 \\
        TemporalFeatrueRL & 0.4223 & 0.2062 & 0.4111 & 0.2109 & 0.2926 & 0.1856 & 0.3053 & 0.1520 \\
        w/o Cross-temporal & 0.4221 & 0.2124 & 0.4308 & 0.2333 & 0.3515 & 0.2302 & 0.3723 & 0.2104 \\
        w/o Cross-feature & 0.4140 & 0.2070 & 0.4160 & 0.2214 & 0.3079 & 0.1999 & 0.2609 & 0.1255 \\
        w/o Feature-Split\&Scale & 0.4147 & 0.2103 & 0.4018 & 0.2051 & 0.2825 & 0.1812 & 0.3100 & 0.1443 \\
        w/o Prominence & 0.4132 & 0.2076 & 0.4041 & 0.2080 & 0.2809 & 0.1799 & 0.2595 & 0.1240 \\
        w/o Node-scale & 0.4130 & 0.2081 & 0.4015 & 0.2055 & 0.2845 & 0.1844 & 0.2631 & 0.1263 \\
        w/o Feature-scale & 0.4213 & 0.2057 & 0.4083 & 0.2093 & 0.2966 & 0.1877 & 0.2932 & 0.1460 \\
        w/o GSL & 0.4218 & 0.2060 & 0.4090 & 0.2095 & 0.2990 & 0.1907 & 0.2962 & 0.1475\\ \midrule
        GSLI & \textbf{0.4101} & \textbf{0.2051} & \textbf{0.3986} & \textbf{0.2034} & \textbf{0.2720} & \textbf{0.1730} & \textbf{0.2534} & \textbf{0.1202} \\ \bottomrule
        \end{tabular}
    }
    \vspace{-0.1in}
     \caption{Ablation analysis of GSLI with 10\% missing values}
    \label{tab:ablation}
\end{table}

Since these variants involve verifying the role of learning graph structures between different features, we performed ablation experiments on four datasets that recorded multiple features, the results are shown in Table \ref{tab:ablation}.
The experimental results indicate that each component of the GSLI plays a crucial role, especially in learning the common spatial correlation between different features within nodes for graph structure learning, and performing cross-temporal representation learning.
It is worth noting that TemporalGCN consistently performs less than TemporalFeatureSA.
This suggests that the given adjacency matrix between nodes cannot accurately reflect the complex spatial correlations in reality.
This also confirms the necessity of adopting different scales for learning graph structures.
 
\section{Conclusion}
In this work, we design the multi-scale Graph Structure Learning framework for spatial-temporal Imputation (GSLI), addressing the challenges of imputing missing values in spatial-temporal data due to feature heterogeneity and latent common correlation between features among all nodes.
By applying node-scale and feature-scale graph structure learning alongside prominence modeling, GSLI improves the imputation accuracy, as demonstrated across six diverse datasets with real missing values.

\section*{Acknowledgements}
This work is supported in part by the National Natural Science Foundation of China (62302241, 62372252, 72342017, 62306085, 62272250), the Natural Science Foundation of Tianjin (No. 22JCJQJC00150), Shenzhen College Stability Support Plan (GXWD20231130151329002).

\bibliography{aaai25}

\clearpage
\section{Appendix}

\subsection{Notations}
Table \ref{table-notations} lists our frequently used notations.

\subsection{Prominence Modeling}
\label{appendix:prominence}

In this section, we present the intuition for introducing the prominence Modeling in the node-scale and feature-scale graph structures learning process in detail.

We first select 10\% observations of the original DutchWind \cite{Dutch} dataset as the missing values by MCAR mechanism \cite{MCAR}.
Then we use cross-temporal self-attention to capture the temporal dependencies and then use cross-feature self-attention to capture the spatial dependencies, based on the setup of the ``TemporalFeatureRL" scenario in the Ablation Study section.
After that, we use the above architecture to impute missing values.
Furthermore, we extract the attention map obtained from cross-feature self-attention after training.
Note that this setup is consistent with the survey addressed in Figure \ref{fig:motivation}(c) and Figure \ref{fig:motivation}(d) in the Introduction section.

\begin{figure}[t]
    \centering
    \includegraphics[width=\linewidth]{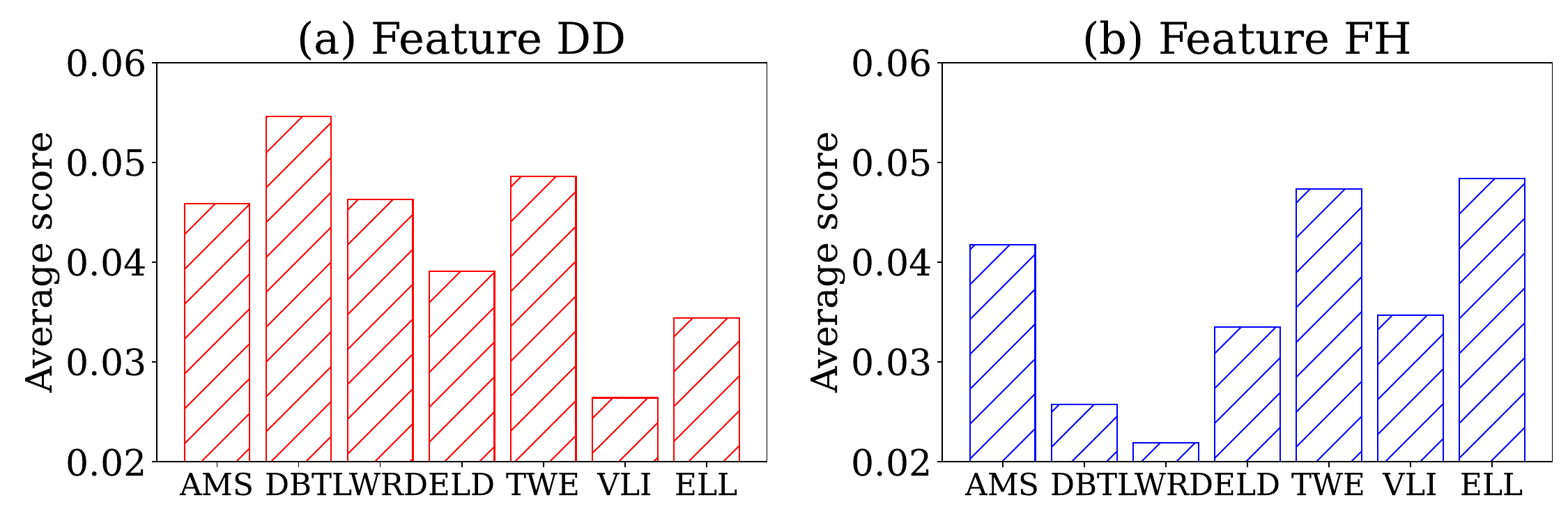}
    \vspace{-0.2in}
    \caption{Average attention scores of different stations from cross-feature self-attention mechanism}
    \label{fig:attn-score-node-scale}
    \vspace{0.3in}

    \centering
    \includegraphics[width=\linewidth]{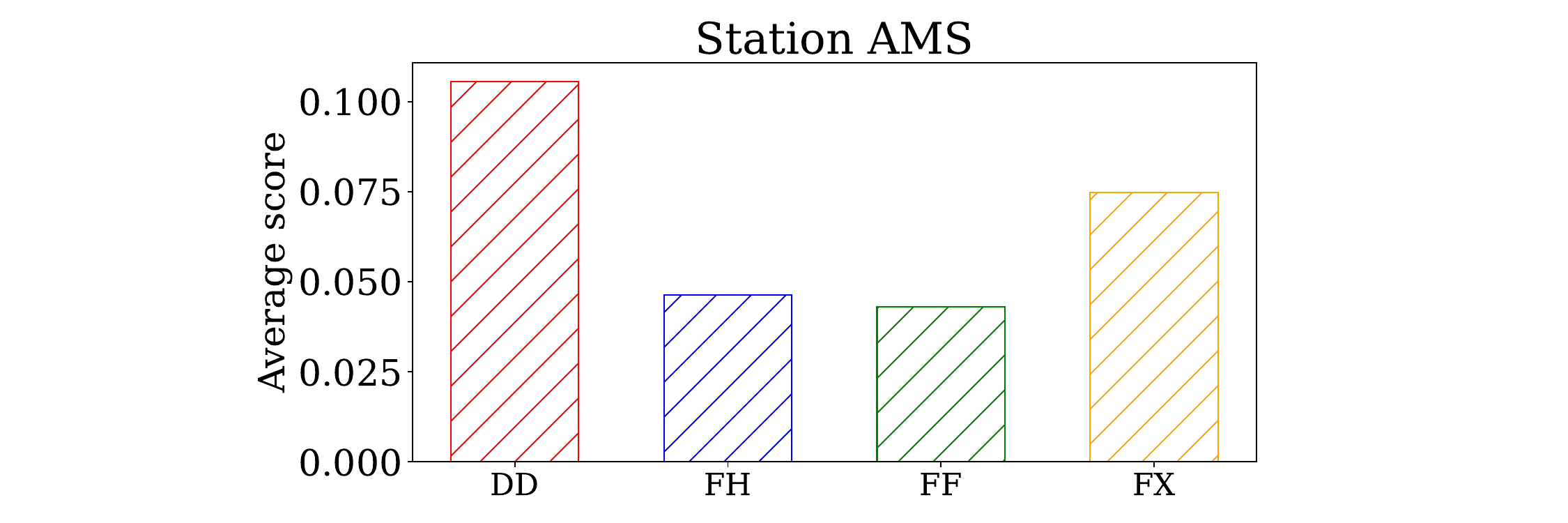}
    \vspace{-0.2in}
    \caption{Average attention scores of different features from cross-feature self-attention mechanism}
    \label{fig:attn-score-feature-scale}
\end{figure}

According to the attention map from the cross-feature self-attention mechanism, we first compute the average attention scores of different stations of each feature.
As shown in Figure \ref{fig:attn-score-node-scale}, the average attention score for different stations is different.
This suggests the influence for the overall imputation task is different for different stations in the node-scale meta-graph corresponding to each feature.

Then we compute the average attention scores of different features in a station.
According to Figure \ref{fig:attn-score-feature-scale}, we can find that the average attention score for different features is different.
Therefore, the influence for the overall imputation task should be different in the feature-scale meta-graph structure.

The above observation inspires us to perform prominence modeling for the node-scale graph structure learning and feature-scale graph structure learning, as presented in Equation \ref{eq:prominence-node}, Equation \ref{eq:refine-node}, Equation \ref{eq:prominence-feature}, and Equation \ref{eq:refine-feature}.
The process first obtains the corresponding prominence vector self-adaptively based on the input source embedding.
And then apply the prominence vector to the source embedding through the Hadamard product to make it enhanced or weakened.

\subsection{Proofs}
\label{proofs}

\begin{table}[t]
\centering
\renewcommand\tabcolsep{4pt}
     \renewcommand{\arraystretch}{1}
     \resizebox{\linewidth}{!}{
     \renewcommand\tabcolsep{2pt}
       \begin{tabular}{rp{2.7in}}
    \toprule  
    Symbol & Description 
    \\
    \midrule
    $\mathbf{X}$ & incomplete temporal signal of spatial-temporal data with $N$ nodes, $T$ timestamps and $F$ features \\
    $\mathcal{G}$ & spatial graph of spatial-temporal data\\
    $\mathbf{M}$ & mask matrix indicating the missing status of temporal signal $\mathbf{X}$\\
    $\mathbf{A}$ & adjacency matrix of the graph $\mathcal{G}$\\
    $\mathbf{R}$ & output of the cross-temporal representation learning module\\
    $\mathbf{R^{NL}}$ & output of the node-scale spatial learning module\\
    $\mathbf{R^{FL}}$ & output of the feature-scale spatial learning module\\
    $\overline{\mathbf{X}}$ & output of GSLI framework\\
    $\mathbf{\Omega}^{\mathbf{1}}_{f}, \mathbf{\Omega}^{\mathbf{2}}_{f}$  & meta node embeddings of feature $f$\\
    $\dot{\mathbf{A}}_{f}^{\mathbf{\Omega}}$ & adjacency matrix of node-scale meta graph $\dot{\mathcal{G}}_{f}^{\mathbf{\Omega}}$ for feature $f$\\
    $\mathbf{\Phi^{1}}, \mathbf{\Phi^{2}}$ & meta feature embeddings across all nodes\\
    $\dot{\mathbf{A}}^{\mathbf{\Phi}}$ & adjacency matrix of feature-scale meta graph $\dot{\mathcal{G}}^{\mathbf{\Phi}}$\\
    \bottomrule
    \end{tabular}
    }
     \caption{Notations}
    \label{table-notations}
\end{table}

\subsubsection{Proof of Proposition \ref{prop:gnn}}
\label{proof-prop-gnn}
The canonical graph diffusion convolution treats all features on a node as a uniform node embedding. Thus, its convolution operation can be expressed as:
\begin{align*}
    \mathbf{R}^{\mathbf{DC}} = \Sigma_{k=0}^{K}
    & \left[ 
    \dot{\mathbf{A}}^{\mathbf{\Omega}} \mathbf{R} \mathbf{\Theta}_{k}^{\mathbf{\Omega 1}} + {\left({\mathbf{D}^{\mathbf{O}}}^{-1} \mathbf{A}\right)}^k \mathbf{R} \mathbf{\Theta}_{k}^{\mathbf{\Omega 2}} +  \right. \\ 
    & \left. {\left({\mathbf{D}^{\mathbf{I}}}^{-1} \mathbf{A}^{\top}\right)}^k \mathbf{R} \mathbf{\Theta}_{k}^{\mathbf{\Omega 3}} \right],
\end{align*}
 where $\dot{\mathbf{A}}^{\mathbf{\Omega}} \in \mathbb{R}^{N \times N}$ is the learned global graph structure, $\mathbf{R} \in \mathbb{R}^{N \times T \times FC}$ is the input graph signal. $\mathbf{\Theta}_{k}^{\mathbf{\Omega 1}},\mathbf{\Theta}_{k}^{\mathbf{\Omega 2}},\mathbf{\Theta}_{k}^{\mathbf{\Omega 3}} \in \mathbb{R}^{FC \times FC}$ are graph convolution kernels.

Considering the feature heterogeneity, then there exists at least one node $i$ satisfying the correlation weight of feature $f_1$ from node $j$ to node $i$ is $x$, and the correlation weight of feature $f_2$ from node $j$ to node $i$ is $y$, where $x \neq y$.
The expected result $\mathbf{B}$ obtained by multiplying $\dot{\mathbf{A}}^{\mathbf{\Omega}}$ with $\mathbf{R}$ in the first term of the first order graph diffusion convolution, the corresponding result $b_{i,t,{f_2},c} \in \mathbf{B}$ of the channel $c$ for $f_2$ feature at timestamp $t$ for the node $i$ is given by:
\begin{equation*}
    (a^{\mathbf{\Omega}}_{i1} r_{1,{f_2},c}+\dots + y r_{j,{f_2},c} + \dots + a^{\mathbf{\Omega}}_{iN} r_{N,{f_2},c}),
\end{equation*}
where $a^{\mathbf{\Omega}}_{i1}, a^{\mathbf{\Omega}}_{iN},\in \dot{\mathbf{A}}^{\mathbf{\Omega}}$, 
$r_{1,{f_2},c}, r_{N,{f_2},c},  r_{j,{f_2},c}\in \mathbf{R}$,
$y$ is the expected value for $a^{\mathbf{\Omega}}_{ij} \in \dot{\mathbf{A}}^{\mathbf{\Omega}}$.

Since the canonical graph diffusion convolution operation has only one learned graph structure, 
if we assume that $\dot{\mathbf{A}}^{\mathbf{\Omega}}$ satisfies the global correlation for $f_1$,
then the actual information flow to form the $b_{i,t,{f_2},c} \in \mathbf{B}$ is:
\begin{equation*}
    b_{i,t,{f_2},c} = 
    a^{\mathbf{\Omega}}_{i1} r_{1,{f_2},c}+\dots + x r_{j,{f_2},c} + \dots + a^{\mathbf{\Omega}}_{iN} r_{N,{f_2},c},
\end{equation*}
which is in conflict with our expected result.
Similarly, if we assume that $\dot{\mathbf{A}}^{\mathbf{\Omega}}$ satisfies the global correlation for $f_2$, the information flow to form the $b_{i,t,{f_1},c} \in \mathbf{B}$ will conflict with our expected result.
Following the same line, we can show that neither the second nor the third term of the graph diffusion convolution addresses the feature heterogeneity.

\subsubsection{Proof of Proposition \ref{prop:gsli}}
\label{proof-prop-gsli}
Similar to Proposition \ref{prop:gnn}, we assume that there exists at least one node $i$ satisfying the correlation weight of feature $f_1$ from node $j$ to node $i$ is $x$, and the correlation weight of feature $f_2$ from node $j$ to node $i$ is $y$ for the feature heterogeneity problem, where $x \neq y$.

According to Equation \ref{eq:node-scale-graph-learning}, we learn independent graph structures for each heterogeneous feature.
Thus, we suppose that $a^{\mathbf{\Omega}}_{f_1,i,j}=x$, $a^{\mathbf{\Omega}}_{f_2,i,j}=y$, where $a^{\mathbf{\Omega}}_{f_1,i,j} \in \dot{\mathbf{A}}^{\mathbf{\Omega}}_{f_1}$ and $a^{\mathbf{\Omega}}_{f_2,i,j} \in \dot{\mathbf{A}}^{\mathbf{\Omega}}_{f_2}$.
If the first term of the first order graph convolution operation shown in Equation \ref{eq:node-scale-graph-learning} operates on the channel $c$ of feature $f_2$, the result obtained by multiplying $\dot{\mathbf{A}}^{\mathbf{\Omega}}_{f_2}$ with $\mathbf{R}_{f2}$ is $\mathbf{B}_{f2}$. 
The corresponding result of $\mathbf{B}_{f_2}$ at timestamp $t$ for the node $i$ can be obtained by:
\begin{equation*}
    b_{f_2,i,t}= a^{\mathbf{\Omega}}_{f_2,i,1} r_{f_2, 1}+ \dots + y r_{f_2,j} + \dots + a^{\mathbf{\Omega}}_{f_2,i,N} r_{f_2, N}.
\end{equation*}

On the other hand, the corresponding result of feature $f_1$ at timestamp $t$ for the node $i$ can be obtained by:
\begin{equation*}
    b_{f_1,i,t}= a^{\mathbf{\Omega}}_{f_1,i,1} r_{f_1, 1}+ \dots + x r_{f_1,j} + \dots + a^{\mathbf{\Omega}}_{f_1,i,N} r_{f_1, N}.
\end{equation*}
It can be seen that the heterogeneity of $f_1$ and $f_2$ does not affect the correct information flow in our node-scale spatial learning process for the timestamp $t$.
According to the above process, we can generalize the above result for all heterogeneous feature pairs and all timestamps.

\subsection{Complexity Analysis}
\label{sec-complex-analysis}
\subsubsection{Time complexity of Node-scale Spatial Learning}
Consider the node-scale spatial learning for feature $f$, the complexity to get the prominence vector ${\mathbf{P}}_{f}^{\mathbf{\Omega}}$ is $\mathcal{O}(Nd^2)$. Then the complexity to obtain the meta-graph $\dot{\mathcal{G}}_{f}^{\mathbf{\Omega}}$ is $\mathcal{O}(N^2d+Nd^2)$. For the graph diffusion convolution shown in Equation \ref{eq:node-scale-graph-conv}, the complexity of the three convolution terms is both $\mathcal{O}(N^2TC+NTC^2)$. Since the graph diffusion convolution step $K$ is a smaller hyperparameter, the overall time complexity of Equation \ref{eq:node-scale-graph-conv} is $\mathcal{O}(N^2TC+NTC^2+N^2d+Nd^2)$.
Finally, Equation \ref{eq:concat-node-scale} concatenates the results obtained on the $F$ features, so the overall time complexity is $\mathcal{O}(FN^2TC+FNTC^2+FN^2d+FNd^2)$.

\subsubsection{Space complexity of Node-scale Spatial Learning}
For the spatial learning on feature $f$, the complexity of Equation \ref{eq:node-scale-graph-learning} is $\mathcal{O}(Nd+d^2)$, thus the complexity of obtaining the meta-graph is $\mathcal{O}(N^2+Nd+d^2)$. And the graph diffusion convolution brings a complexity of $\mathcal{O}(NTC+C^2)$.
Since we need to concatenate all the features, the overall space complexity is $\mathcal{O}(FNTC+FC^2+FN^2+FNd+Fd^2)$.

\subsubsection{Time complexity of Feature-scale Spatial Learning}
First, we need to get the adjacency matrix $\dot{\mathbf{A}}^{\mathbf{\Phi}}$ of the meta-graph which represents the spatial correlation of different features, the time complexity of this operation is $\mathcal{O}(F^2d+Fd^2)$. Then the time complexity of the diffusion convolution layer shown in Equation 10 is $\mathcal{O}(F^2NTC+FNTC^2)$. Therefore, the overall time complexity is $\mathcal{O}(F^2NTC+FNTC^2+F^2d+Fd^2)$.

\subsubsection{Space complexity of Feature-scale Spatial Learning}
Similar to the node-scale spatial learning, the complexity of obtaining the meta-graph $\dot{\mathcal{G}}^{\mathbf{\Phi}}$ is $\mathcal{O}(F^2+Fd+d^2)$. The space complexity of graph diffusion convolution is $\mathcal{O}(FNTC+C^2)$. The overall space complexity is $\mathcal{O}(FNTC+C^2+F^2+Fd+d^2)$.

\begin{figure}[t]
    \centering
    \includegraphics[width=\linewidth]{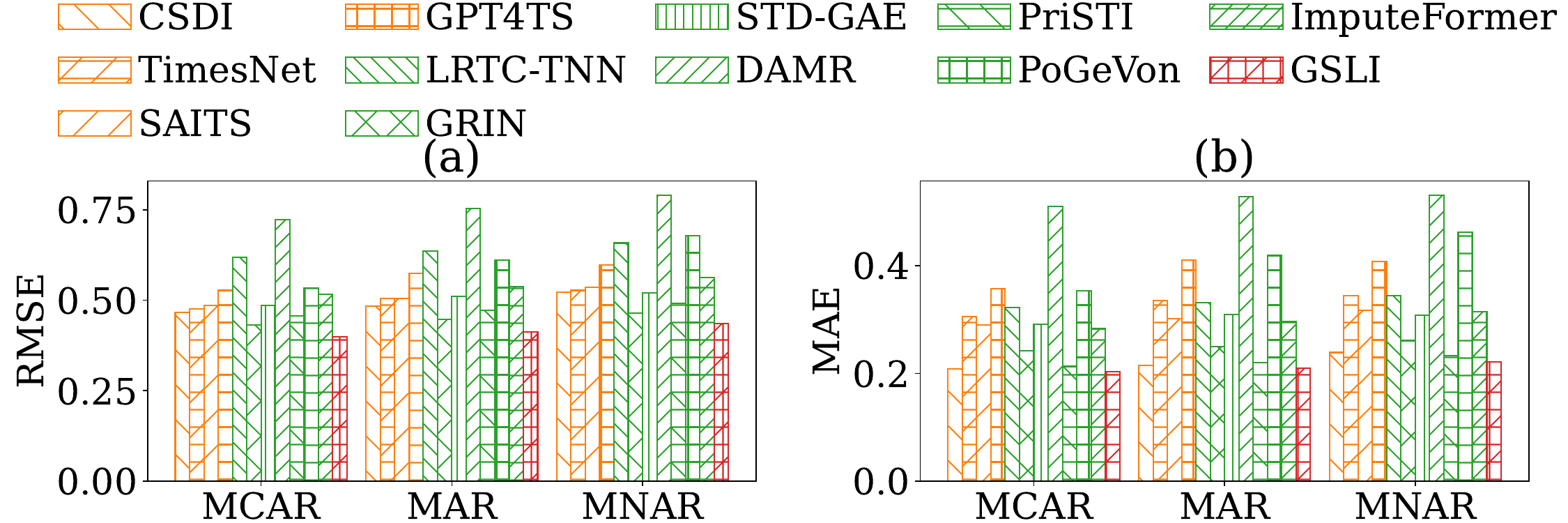}
    \caption{Varying the missing mechanism over BeijingMEO dataset with 10\% missing values}
    \label{fig:missingmechanism_beijingmeo}
    \vspace{0.3in}

    \includegraphics[width=\linewidth]{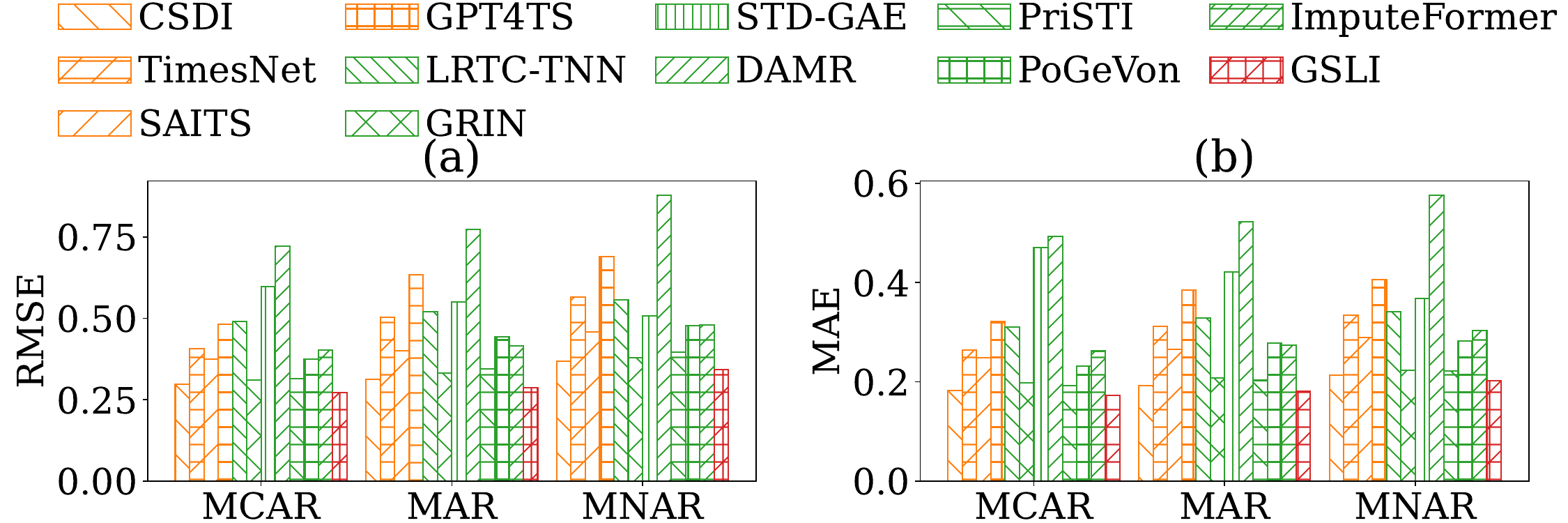}
    \caption{Varying the missing mechanism over LondonAQ dataset with 10\% missing values}
    \label{fig:missingmechanism_london}
    \vspace{0.3in}

    \includegraphics[width=\linewidth]{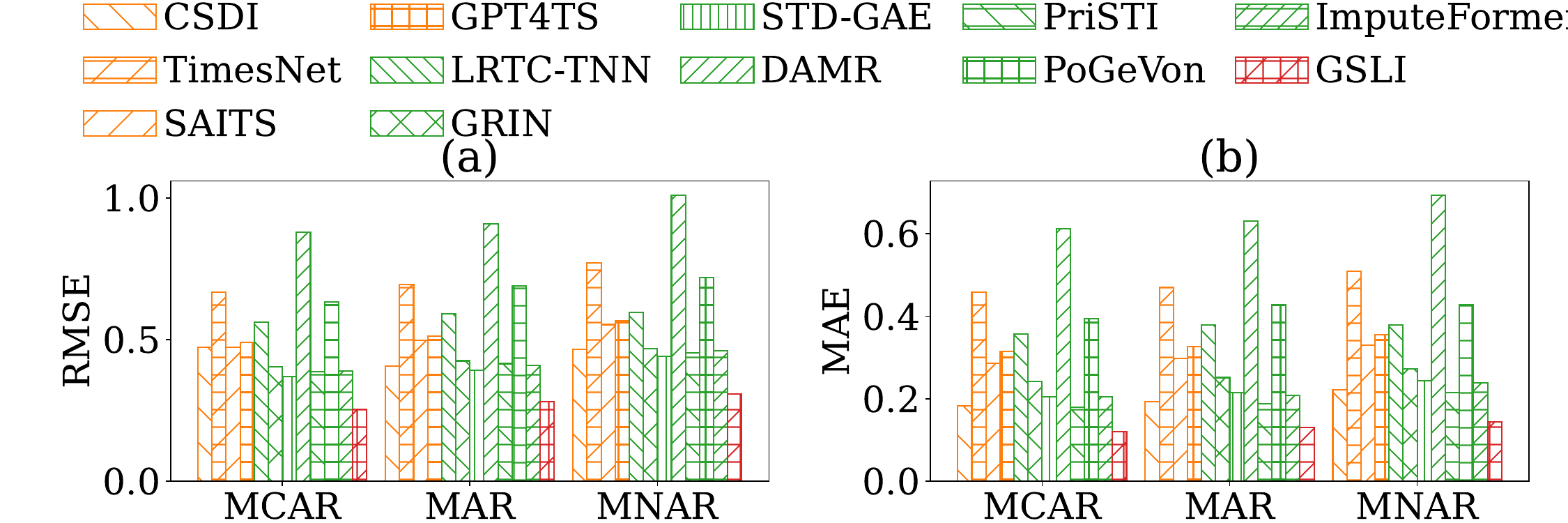}
    \caption{Varying the missing mechanism over CN dataset with 10\% missing values}
    \label{fig:missingmechanism_cn}
    \vspace{0.3in}
\end{figure}

\begin{figure}[t]
    \includegraphics[width=\linewidth]{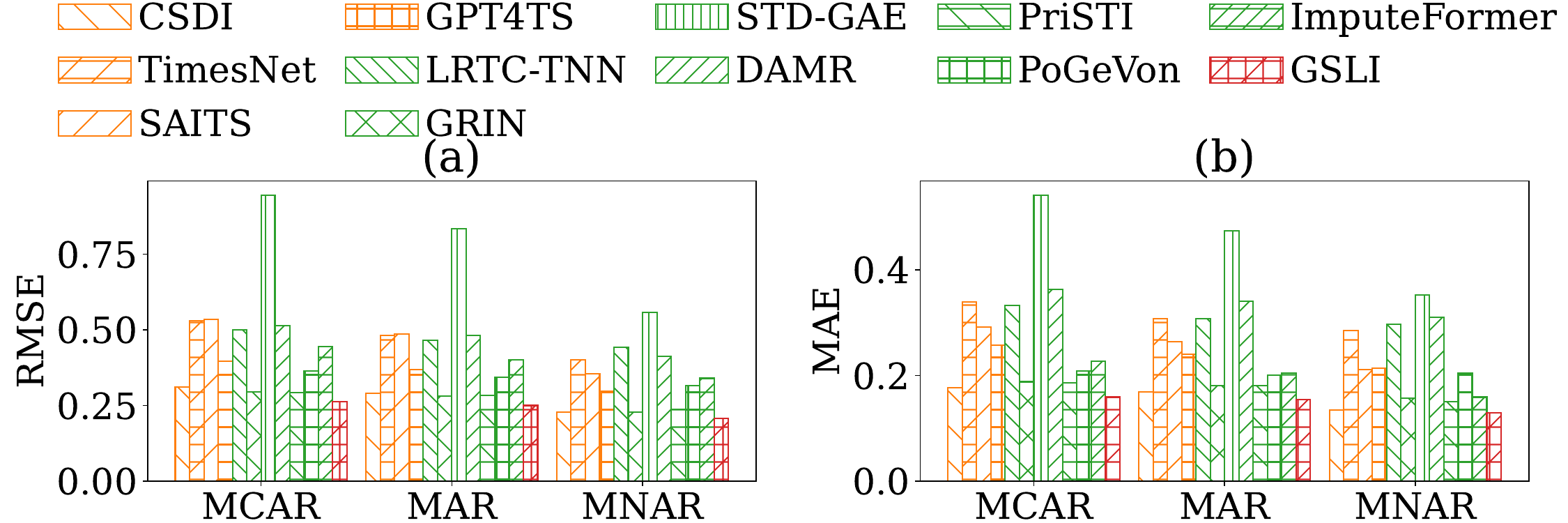}
    \caption{Varying the missing mechanism over Los dataset with 10\% missing values}
    \label{fig:missingmechanism_los}
    \vspace{0.3in}
    
   \includegraphics[width=\linewidth]{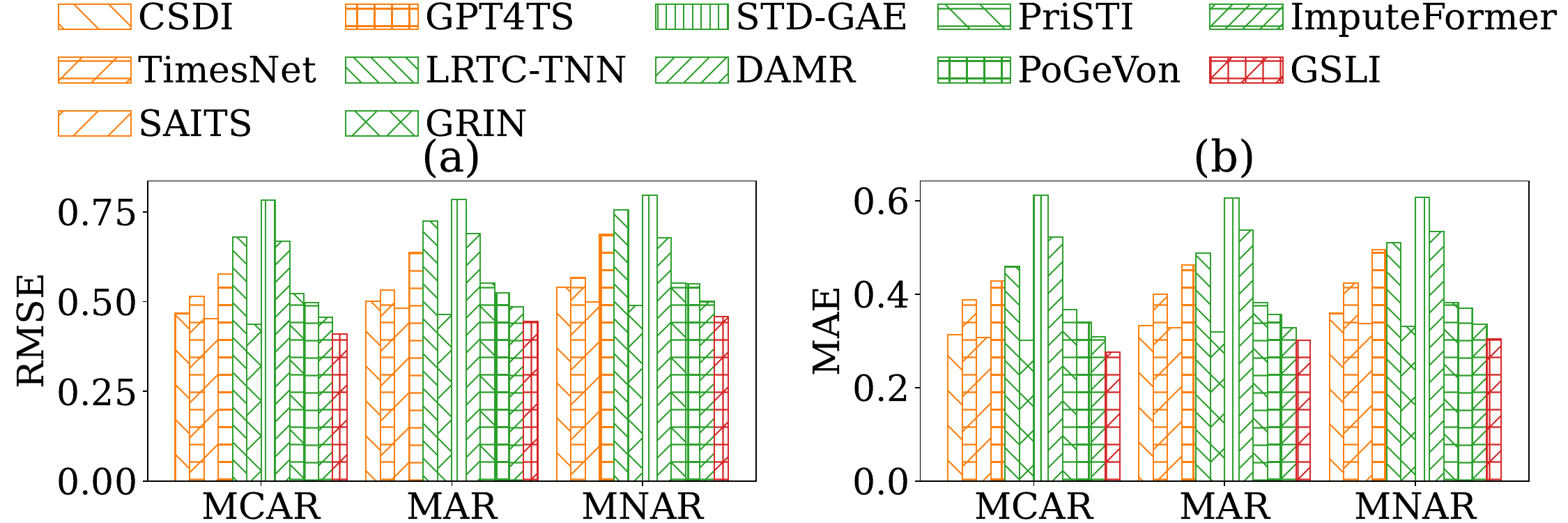}
    \caption{Varying the missing mechanism over LuohuTaxi dataset with 10\% missing values}
    \label{fig:missingmechanism_luohutaxi}
\end{figure}

\subsection{Supplementary of experiments}
\subsubsection{Dataset Details}
DutchWind \cite{Dutch} gathers wind speed and direction data from 7 stations in the Netherlands hourly from January 1 to December 28, 2023.
BeijingMEO \cite{Beijing} contains meteorological data recorded at 18 locations in Beijing from January 30, 2017, to January 31, 2018, collected hourly.
LondonAQ  \cite{London} collects hourly air quality readings from January 1, 2017 to March 31, 2018 at 13 locations in London.
CN \cite{CN} contains hourly air quality data at 140 stations in China from October 1, 2014, to December. 31, 2014. Each station records six features: PM2.5, PM10, NO2, CO, O3, and SO2.
Los \cite{Traffic} contains average traffic speeds per five-minute period from March 1, 2012 to March 7, 2012 at different locations on Los Angeles highways.
LuohuTaxi \cite{Traffic} records the average speed of taxis every 15 minutes on different major roads in Luohu District, Shenzhen in January 2015.

\begin{table*}[t]
\centering
\renewcommand\tabcolsep{4pt}
\begin{tabular}{@{}l|ll|ll|ll|ll|ll|ll@{}}
\toprule
Mask Ratio & \multicolumn{2}{l|}{DutchWind} & \multicolumn{2}{l|}{BeijingMEO} & \multicolumn{2}{l|}{LondonAQ} & \multicolumn{2}{l|}{CN} & \multicolumn{2}{l|}{Los} & \multicolumn{2}{l}{LuohuTaxi} \\ \cmidrule(l){2-13} 
 & RMSE & MAE & RMSE & MAE & RMSE & MAE & RMSE & MAE & RMSE & MAE & RMSE & MAE \\ \midrule
10\% & 0.4143 & 0.2080 & 0.4114 & 0.2153 & 0.3034 & 0.1963 & 0.2582 & 0.1233 & 0.2697 & 0.1622 & 0.4186 & 0.2841 \\
20\% & \textbf{0.4101} & 0.2051 & \textbf{0.3986} & \textbf{0.2034} & 0.2720 & 0.1730 & 0.2534 & \textbf{0.1202} & \textbf{0.2632} & \textbf{0.1592} & \textbf{0.4102} & \textbf{0.2761} \\
30\% & 0.4114 & \textbf{0.2038} & \textbf{0.3986} & \textbf{0.2034} & 0.2718 & 0.1731 & \textbf{0.2520} & 0.1203 & 0.2635 & 0.1612 & 0.4147 & 0.2819 \\
40\% & 0.4226 & 0.2124 & 0.3995 & 0.2051 & \textbf{0.2708} & \textbf{0.1728} & 0.2530 & 0.1207 & 0.2651 & 0.1616 & 0.4174 & 0.2836 \\
50\% & 0.4318 & 0.2179 & 0.4022 & 0.2142 & 0.2800 & 0.1752 & 0.2563 & 0.1227 & 0.2704 & 0.1649 & 0.4183  & 0.2851   \\ \bottomrule
\end{tabular}
\vspace{-0.1in}
\caption{Varying the mask ratio of the training label for various datasets with 10\% missing values}
    
    \vspace{-0.1in}
    \label{tab:mask-ratio}
\end{table*}

\subsubsection{Missing Mechanisms}
\label{appendix-missing-mechanism}

In this section, we provide the imputation performance with various missing mechanisms for all datasets.

The occurrence of missing data in real-world scenarios is usually related to the external environment or the sensors themselves.
Therefore, we consider three missing mechanisms: missing completely at random (MCAR) \cite{MCAR}, missing at random (MAR) \cite{MAR} and missing not at random (MNAR) \cite{MNAR}.
For MCAR, the missing value is not related to other attributes, and each observation has an equal chance of being missing.
MAR potentially implies that the missing status of all features depends on the frequency of a particular feature. For example, missing records of traffic flow data may be related to rush hour and activities in public places.
For MNAR, the missing observation depends on the feature itself, for example, the reliability of recording devices.
As in the Imputation Comparison section, we repeat each experiment 5 times for each experiment and report the average results.
It should be noted that in all experiments we used different random seeds in our 5 replications, i.e. from 3407 to 3411.
Since the random seeds are different, the imputation labels selected in the five replications will also be different.

As shown in Figure \ref{fig:missingmechanism_dutch}, Figure \ref{fig:missingmechanism_beijingmeo}, Figure \ref{fig:missingmechanism_london}, Figure \ref{fig:missingmechanism_cn}, Figure \ref{fig:missingmechanism_los}, and Figure \ref{fig:missingmechanism_luohutaxi},  our GSLI method consistently delivers optimal results across different missing mechanisms.
It demonstrates that our proposed GSLI is effectively adapted to various missing scenarios in reality.
Furthermore, it is important to note that when it comes to the MNAR mechanism, most imputation methods tend to perform slightly worse than the other mechanisms across the majority of datasets.
This is because MNAR tends to remove observations in unconventional statuses.

\subsubsection{Mask Strategies of Training Label}
For training our GSLI framework, we randomly select some observations as the training label.
In this process, the mask ratio and mask pattern to get the training label directly determine the effectiveness of training.
Thus, in this section, we explore the performance with different mask ratios and mask patterns.

We first mask different ratios of observations as the training label, the performance of GSLI with different mask ratios is shown in Table \ref{tab:mask-ratio}.
We can find that the model performs better when the mask ratio is set to 0.2 or 0.3 in most cases.
The training will be easy to converge when the mask ratio is too low, and therefore the model will not be able to accurately learn the the required dependencies for imputation.
On the contrary, when the mask ratio is too large, it is difficult for the model to mine valid dependencies from training.
Therefore, we set the mask ratio to 0.2 by default.

Then we explore the performance when using different mask patterns.
Following CSDI \cite{CSDI} and PriSTI \cite{PRISTI}, we consider three mask pattern strategies: (1) Block missing (2) Historical missing (3) Random missing.
For the ``Historical missing" scenario, the mask at the current timestamp has a half probability of being the same as the previous timestamp, and a half probability of performing Random missing.
As shown in Table \ref{tab:mask-pattern}, the Block missing or Historical missing strategy is not directly comparable to Random missing in most cases. This is because there is a possibility that the mask pattern may not accurately correspond to the actual missing scenario. Therefore, we default to using the Random missing mask pattern.
It should be noted that by comparing the results presented in Table \ref{tab:compare-exp}, it can be found that benefiting from the learning of multi-scale graph structures, our method consistently outperforms other methods when adjusting the mask strategies in all of them.

\begin{table*}[h]
\centering
\vspace{0.1in}
\renewcommand\tabcolsep{4pt}
\begin{tabular}{@{}l|ll|ll|ll|ll|ll|ll@{}}
\toprule
Mask Pattern & \multicolumn{2}{l|}{DutchWind} & \multicolumn{2}{l|}{BeijingMEO} & \multicolumn{2}{l|}{LondonAQ} & \multicolumn{2}{l|}{CN} & \multicolumn{2}{l|}{Los} & \multicolumn{2}{l}{LuohuTaxi} \\ \cmidrule(l){2-13}  
 & RMSE & MAE & RMSE & MAE & RMSE & MAE & RMSE & MAE & RMSE & MAE & RMSE & MAE \\ \midrule
Block missing & 0.4150 & 0.2057 & 0.4089 & 0.2117 & 0.3328 & 0.2197 & 0.2887 & 0.1447 & 0.2759 & 0.1674 & 0.4748 & 0.3209 \\
Historical missing & \textbf{0.4101} & \textbf{0.2041} & 0.4013 & 0.2057 & 0.3068 & 0.1989 & 0.2786 & 0.1385 & 0.2636 & 0.1596 & 0.4621 & 0.3080 \\
Random misssing & \textbf{0.4101} & 0.2051 & \textbf{0.3986} & \textbf{0.2034} & \textbf{0.2720} & \textbf{0.1730} & \textbf{0.2534} & \textbf{0.1202} & \textbf{0.2632} & \textbf{0.1592} & \textbf{0.4102} & \textbf{0.2761} \\ \bottomrule
\end{tabular}
 \vspace{-0.1in}
\caption{Varying the mask pattern of the training label for various datasets with 10\% missing values}
    \label{tab:mask-pattern}
\end{table*}
\begin{figure*}[h]
    \centering
    \includegraphics[width=0.8\linewidth]{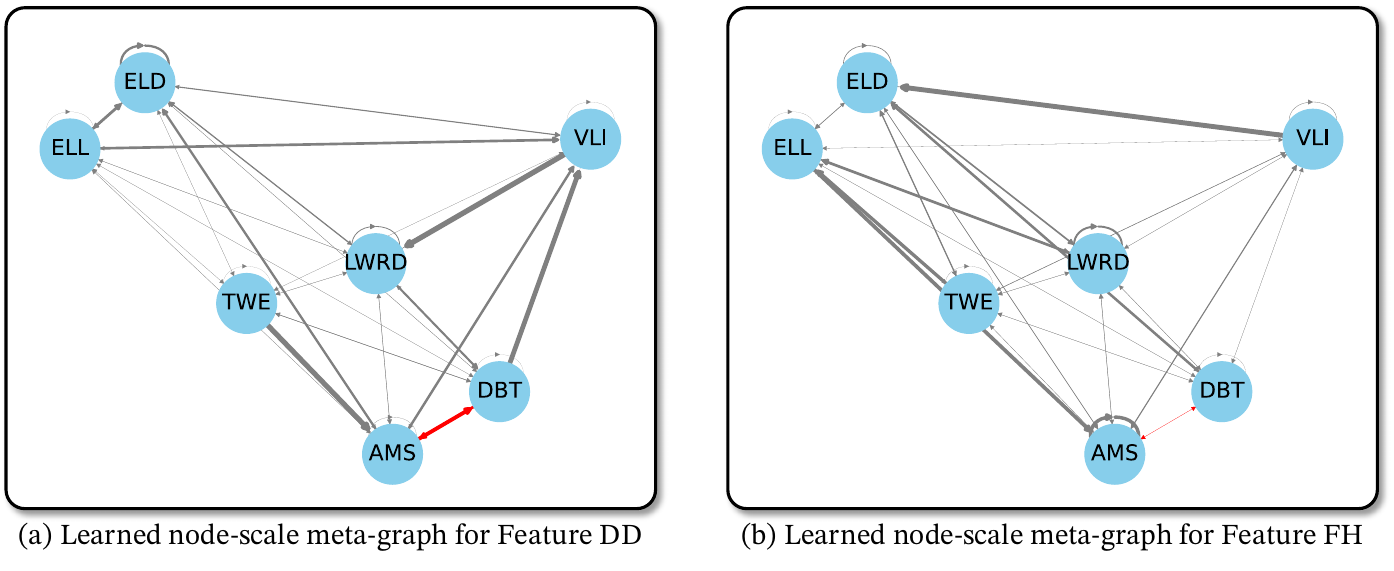}
    \vspace{-0.2in}
    \caption{Visualisation of the learned node-scale meta-graphs}
    \label{fig:node-scale-graph}
\end{figure*}

\subsubsection{Resource Consumption}
\label{appendix-resource}

\begin{table}[H]
\centering
\renewcommand\tabcolsep{4pt}
     \resizebox{\linewidth}{!}{
     \begin{tabular}{@{}llll@{}}
    \toprule
    Methods & Parameters & \begin{tabular}[c]{@{}l@{}}GPU Memory Usage\\(MiB)\end{tabular} &  Time Cost(s) \\ \midrule
    CSDI & 413441 & 1888 & 87.78 \\
    TimesNet & 713692 & 604 & 26.25 \\
    SAITS & 1362896 & 1044 & 1.42 \\
    GPT4TS & 60736540 & 1386 & 89.11 \\
    LRTC-TNN & - & - & 4.96 \\
    GRIN & 24101 & 1950 & 126.01 \\
    STD-GAE & 3803 & 432 & 212.38 \\
    DAMR & 2977 & 23080 & 5518.04 \\
    PriSTI & 729202 & 2266 & 1129.58 \\ 
    PoGeVon & 329616 & 2868 & 67.09 \\ 
    ImputeFormer & 798585 & 2934 & 23.70 \\ 
    \midrule
    w/o Feature-Split\&Scale & 4281781 & 2354 & 47.70 \\
    GSLI & 4478913 & 2464 & 47.90 \\ \bottomrule
    \end{tabular}
    }
    \caption{Resource consumption over DutchWind dataset with 10\% missing values}
    \label{tab:resource}
\end{table}
In this section, we report the resource consumption in Table \ref{tab:resource} for our GSLI and existing methods.
In this table, we also consider another ablation scenario ``w/o FeatureSplit\&Feature-scale".
As described in Ablation Study Section, we replace our node-scale Spatial Learning and feature-scale Spatial Learning with the canonical graph diffusion convolution \cite{DCRNN,GWN}.

We can find that the overall resource consumption of our method is in the same order of magnitude as the end-to-end deep learning imputation methods.
Note that LRTC-TNN is based on low-rank tensor completion with no learnable parameters and no need to utilize GPU training.
Since the canonical graph diffusion convolution treats all features on a node as a uniform node embedding, the time complexity is $\mathcal{O}(FN^2TC+F^NTC^2+Nd^2)$, the space complexity is $\mathcal{O}(FNTC+F^2C^2)$. Thus, our resource consumption is slightly higher than the canonical graph diffusion convolution.
However, as illustrated in Table \ref{tab:ablation}, our method can produce better imputation results than the above ablation scenario.
Therefore, we believe that it is acceptable to pay a small amount of additional resource consumption.

\subsubsection{Graph Structure Visualisation}
In this section, we visualize the graph structures we have learned over the DutchWind dataset with 10\% missing rate.

First, we present the learned node-scale meta-graphs for heterogeneous features from our proposed GSLI model, as illustrated in Figure \ref{fig:node-scale-graph}.
In the presented graphs, thicker edges represent higher edge weights, i.e., more spatial correlations are learned.
As mentioned in the Introduction section, feature DD captures wind direction data and FH is related to wind speed.
As shown in Figure \ref{fig:motivation}(a), AMS and DBT stations are geographically close to each other.
Therefore, there is a stronger spatial correlation between the nodes corresponding to AMS and DBT in the node-scale meta-graph corresponding to the feature DD that records the wind direction.
In contrast, due to the large difference in emptiness degrees in the vicinity of AMS and DBT stations, the node-scale meta-graph corresponding to the feature FH for recording wind speeds does not have a strong correlation between the two nodes.
This result shows that our GSLI can learn different global node-scale graph structures for features from different domains in response to feature heterogeneity, which also provides an empirical evidence for Proposition \ref{prop:gnn}.

\begin{figure}[t]
    \centering
    \includegraphics[width=0.8\linewidth]{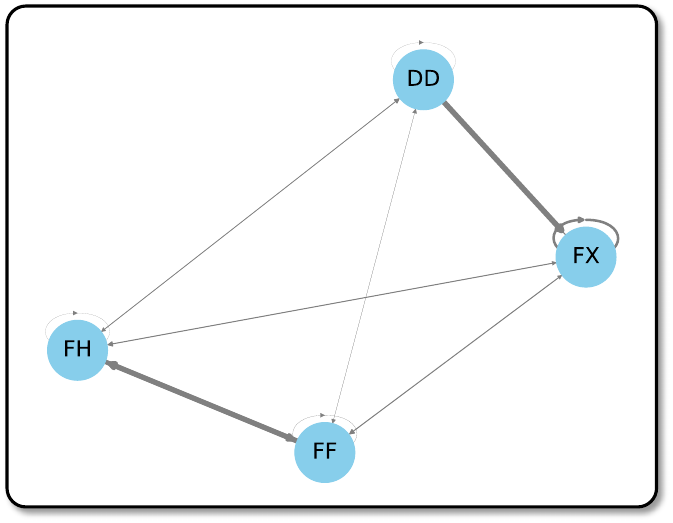}
    \vspace{-0.1in}
    \caption{Visualisation of the learned feature-scale meta-graph}
    \label{fig:feature-scale-graph}
\end{figure}

Then we present the learned feature-scale meta-graph, which represents the common spatial correlation of different features over all nodes learned by GSLI.
As shown in Figure \ref{fig:feature-scale-graph}, there is a stronger spatial correlation between FH and FF in the feature-scale meta-graph.
Given the strong correlation between features FH and FF, i.e. FH records the hourly average wind speed and FF records the average wind speed in the last 10 minutes of the past hour, it can be shown that GSLI can capture the common spatial correlations of different features over all nodes through the feature-scale graph structure learning.

\begin{table}[t]
\centering
\renewcommand\tabcolsep{4pt}
    \resizebox{\linewidth}{!}{
     \renewcommand\tabcolsep{14pt}
        \begin{tabular}{@{}l|ll|ll@{}}
        \toprule
        Method & \multicolumn{2}{l|}{LondonAQ} & \multicolumn{2}{l}{LuohuTaxi} \\ \cmidrule(l){2-5} 
         & RMSE & MAE & RMSE & MAE \\ \midrule
        TimesNet & 0.886 & 0.648 & 0.372 & 0.272 \\
        GPT4TS & 0.885 & 0.666 & 0.371 & 0.266 \\
        DCRNN & 0.948 & 0.738 & 0.764 & 0.596 \\
        GWN & 0.967 & 0.739 & 0.402 & 0.300 \\
        GTS & 0.866 & 0.665 & 0.490 & 0.377 \\
        MegaCRN & 1.075 & 0.783 & 0.373 & 0.274 \\
        CrossGNN & 1.052 & 0.736 & 0.566 & 0.423 \\
        GSLI & \textbf{0.806} & \textbf{0.600} & \textbf{0.368} & \textbf{0.267} \\ \bottomrule
        \end{tabular}
    }
     \caption{Forecasting performance of GSLI compared to existing methods over real missing datasets}
    \label{tab:forecasting}
\end{table}

\subsubsection{Application Study}
\label{sec:exp-applocation}
\begin{figure}[t]
    \centering
    \includegraphics[width=\linewidth]{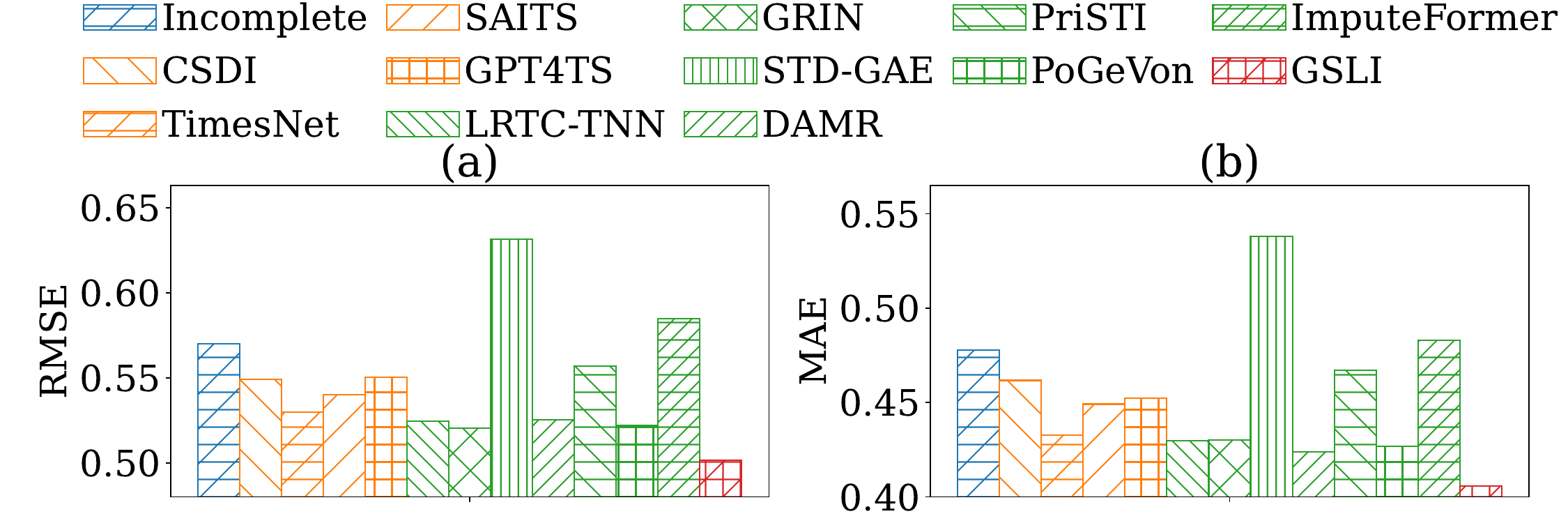}
    \vspace{-0.3in}
    \caption{Downstream application results of air quality forecasting over LondonAQ dataset}
    \label{fig:application}
    \vspace{-0.25in}
\end{figure}

In this section, we validate different imputation methods for the downstream tasks on the original incomplete LondonAQ dataset.
We first impute the missing values through various imputation methods.
Then, we evaluate the accuracy of air quality forecasts following the same line of existing study \cite{GAN19}.
To be more specific, we use the Adaboost implementation \cite{sklearn} to forecast the average PM2.5 concentration at the CD1 station for the next six hours data according to the data from all stations during a 12-hour period.
According to Figure \ref{fig:application}, most methods with higher imputation accuracies tend to have better air quality forecasting performance. 
Meanwhile, our method still obtains the best forecasting result, validating its applicability.

\subsubsection{Spatial-Temporal Forecasting}

In this section, we directly apply GSLI for downstream spatial-temporal forecasting task on datasets with missing values. We utilize 96 historical timestamps to predict 48 future timestamps for the LondonAQ and LuohuTaxi datasets. To meet the prediction requirements, we introduce an additional linear layer that adjusts the output dimensions to match the forecasting length.

As demonstrated in Table \ref{tab:forecasting}, GSLI outperforms existing state-of-the-art methods for spatial-temporal forecasting in situations with missing data. This advantage is due to GSLI's ability to model complex spatial relationships and adapt its learning of node-scale and feature-scale graph structures to account for feature heterogeneity. 
In missing scenarios, temporal dependencies are modelled less accurately, but other methods cannot accurately model spatial dependencies, which affects their forecasting performance.
The results confirm GSLI's superior performance and highlight its robustness in spatial-temporal forecasting when dealing with missing data scenarios.


\subsubsection{Statistical Analysis}
In this section, we first conduct a t-test based on the experimental results shown in Table \ref{tab:compare-exp}.
As shown in Table \ref{tab:t-test-exp}, GSLI performs significantly better than various baselines in $521/528\approx98.77\%$ cases under the t-test threshold ($p<0.05$).
Then we report the the average improvement percentage of  the imputation performance between existing methods and GSLI based on Table \ref{tab:compare-exp}, and the results are shown in Table \ref{tab:improve-percentage-exp}.
The results show that GSLI shows at least an average improvement of \textbf{10.81\%} in performance metrics compared to the second-best results.

\begin{table*}[h]
\centering
     \renewcommand\tabcolsep{1pt}
     \renewcommand{\arraystretch}{1}
     \resizebox{\linewidth}{!}{
     \renewcommand\tabcolsep{4pt}
\begin{tabular}{@{}lll|lllllllllllll@{}}
\toprule
Dataset & Missing rate & Metric & CSDI & TimesNet & SAITS & GPT4TS & LRTC-TNN & GRIN & STD-GAE & DAMR & PriSTI & PoGeVon & ImputeFormer & GSLI \\ \midrule
DutchWind & 10\% & RMSE & \textbf{3.4E-05} & \textbf{4.1E-09} & \textbf{7.4E-07} & \textbf{1.0E-10} & \textbf{2.2E-08} & \textbf{1.6E-04} & \textbf{5.4E-07} & \textbf{4.8E-10} & \textbf{6.6E-07} & \textbf{1.3E-02} & \textbf{2.4E-08} & - \\
 &  & MAE & \textbf{4.5E-05} & \textbf{2.2E-12} & \textbf{3.0E-08} & \textbf{2.7E-13} & \textbf{1.2E-10} & \textbf{8.2E-08} & \textbf{3.5E-10} & \textbf{3.7E-10} & \textbf{5.5E-05} & \textbf{3.0E-02} & \textbf{1.0E-10} & - \\
 & 20\% & RMSE & \textbf{1.6E-04} & \textbf{3.9E-11} & \textbf{2.1E-05} & \textbf{1.5E-12} & \textbf{3.6E-09} & \textbf{7.4E-04} & \textbf{1.4E-07} & \textbf{1.6E-08} & \textbf{1.1E-07} & \textbf{1.5E-02} & \textbf{3.2E-07} & - \\
 &  & MAE & \textbf{3.2E-04} & \textbf{2.7E-14} & \textbf{1.7E-07} & \textbf{8.2E-17} & \textbf{2.7E-11} & \textbf{2.5E-08} & \textbf{6.8E-12} & \textbf{3.8E-10} & \textbf{2.4E-05} & \textbf{1.6E-02} & \textbf{5.8E-09} & - \\
 & 30\% & RMSE & \textbf{8.3E-06} & \textbf{1.3E-12} & \textbf{1.2E-06} & \textbf{1.3E-12} & \textbf{5.2E-11} & \textbf{1.2E-02} & \textbf{2.6E-08} & \textbf{1.3E-08} & \textbf{1.9E-06} & \textbf{3.8E-02} & \textbf{1.5E-07} & - \\
 &  & MAE & \textbf{1.0E-04} & \textbf{1.4E-15} & \textbf{1.1E-10} & \textbf{1.0E-15} & \textbf{4.6E-11} & \textbf{2.8E-06} & \textbf{3.6E-11} & \textbf{2.4E-11} & \textbf{3.4E-05} & \textbf{4.4E-02} & \textbf{1.3E-09} & - \\
 & 40\% & RMSE & \textbf{2.9E-04} & \textbf{5.0E-13} & \textbf{1.8E-06} & \textbf{4.0E-13} & \textbf{2.8E-11} & 1.5E-01 & \textbf{4.8E-09} & \textbf{3.4E-08} & \textbf{6.5E-05} & 7.1E-02 & \textbf{3.2E-07} & - \\
 &  & MAE & \textbf{8.2E-04} & \textbf{2.8E-14} & \textbf{1.2E-07} & \textbf{1.6E-14} & \textbf{3.7E-11} & \textbf{2.2E-03} & \textbf{3.8E-10} & \textbf{2.2E-10} & \textbf{3.4E-03} & 6.2E-02 & \textbf{1.5E-08} & - \\ \midrule
BeijingMEO & 10\% & RMSE & \textbf{6.5E-08} & \textbf{2.9E-08} & \textbf{8.3E-09} & \textbf{6.6E-10} & \textbf{2.8E-11} & \textbf{1.9E-05} & \textbf{9.3E-09} & \textbf{1.1E-10} & \textbf{6.0E-07} & \textbf{1.0E-07} & \textbf{2.1E-08} & - \\
 &  & MAE & 2.2E-01 & \textbf{1.7E-09} & \textbf{1.4E-08} & \textbf{2.2E-10} & \textbf{8.7E-10} & \textbf{6.2E-06} & \textbf{5.4E-09} & \textbf{5.1E-09} & \textbf{4.7E-02} & \textbf{2.7E-07} & \textbf{9.0E-08} & - \\
 & 20\% & RMSE & \textbf{9.3E-09} & \textbf{2.3E-10} & \textbf{1.8E-09} & \textbf{4.9E-12} & \textbf{8.4E-12} & \textbf{3.8E-06} & \textbf{1.5E-09} & \textbf{3.8E-14} & \textbf{2.1E-07} & \textbf{2.5E-06} & \textbf{9.1E-10} & - \\
 &  & MAE & \textbf{5.7E-02} & \textbf{3.4E-11} & \textbf{6.0E-09} & \textbf{2.3E-12} & \textbf{1.9E-10} & \textbf{3.4E-06} & \textbf{2.8E-09} & \textbf{1.3E-11} & \textbf{4.5E-03} & \textbf{5.4E-05} & \textbf{2.1E-08} & - \\
 & 30\% & RMSE & \textbf{2.1E-08} & \textbf{2.9E-11} & \textbf{1.4E-08} & \textbf{2.0E-12} & \textbf{2.1E-12} & \textbf{2.4E-05} & \textbf{8.0E-09} & \textbf{3.4E-12} & \textbf{9.3E-06} & \textbf{4.3E-06} & \textbf{1.9E-08} & - \\
 &  & MAE & \textbf{2.5E-02} & \textbf{9.3E-12} & \textbf{2.5E-08} & \textbf{9.1E-13} & \textbf{2.3E-10} & \textbf{1.3E-05} & \textbf{1.7E-08} & \textbf{1.3E-10} & \textbf{1.1E-03} & \textbf{5.4E-05} & \textbf{1.0E-07} & - \\
 & 40\% & RMSE & \textbf{8.8E-09} & \textbf{6.9E-13} & \textbf{4.4E-09} & \textbf{1.9E-13} & \textbf{1.7E-13} & \textbf{1.1E-05} & \textbf{2.8E-09} & \textbf{5.2E-11} & \textbf{5.6E-03} & \textbf{4.3E-04} & \textbf{5.6E-08} & - \\
 &  & MAE & \textbf{8.5E-04} & \textbf{2.1E-13} & \textbf{5.3E-09} & \textbf{4.5E-14} & \textbf{1.5E-11} & \textbf{2.0E-06} & \textbf{3.6E-09} & \textbf{8.0E-10} & \textbf{2.5E-04} & \textbf{4.7E-03} & \textbf{5.5E-08} & - \\ \midrule
LondonAQ & 10\% & RMSE & \textbf{2.0E-03} & \textbf{2.1E-08} & \textbf{1.2E-07} & \textbf{2.0E-09} & \textbf{1.5E-07} & \textbf{1.8E-04} & \textbf{1.2E-11} & \textbf{4.7E-10} & \textbf{3.8E-04} & \textbf{8.8E-07} & \textbf{7.3E-08} & - \\
 &  & MAE & \textbf{3.1E-02} & \textbf{4.0E-09} & \textbf{4.3E-08} & \textbf{2.6E-10} & \textbf{3.0E-10} & \textbf{8.7E-05} & \textbf{1.2E-12} & \textbf{3.3E-12} & \textbf{1.3E-03} & \textbf{2.3E-07} & \textbf{2.0E-08} & - \\
 & 20\% & RMSE & 2.1E-01 & \textbf{4.9E-09} & \textbf{4.0E-05} & \textbf{4.3E-10} & \textbf{1.8E-08} & \textbf{3.3E-02} & \textbf{3.6E-10} & \textbf{4.0E-11} & \textbf{1.1E-02} & \textbf{3.5E-05} & \textbf{9.9E-06} & - \\
 &  & MAE & 5.6E-01 & \textbf{4.3E-11} & \textbf{6.5E-07} & \textbf{1.9E-12} & \textbf{1.1E-10} & \textbf{1.2E-03} & \textbf{9.2E-13} & \textbf{1.6E-10} & \textbf{5.7E-03} & \textbf{3.0E-07} & \textbf{1.4E-07} & - \\
 & 30\% & RMSE & \textbf{2.4E-02} & \textbf{7.6E-12} & \textbf{4.8E-07} & \textbf{2.7E-12} & \textbf{2.3E-11} & \textbf{1.7E-03} & \textbf{3.1E-05} & \textbf{2.5E-10} & \textbf{3.8E-02} & \textbf{3.8E-06} & \textbf{7.7E-07} & - \\
 &  & MAE & \textbf{1.2E-01} & \textbf{7.7E-12} & \textbf{1.6E-07} & \textbf{1.5E-12} & \textbf{2.0E-10} & \textbf{1.1E-03} & \textbf{9.1E-07} & \textbf{7.6E-10} & \textbf{7.8E-03} & \textbf{1.9E-06} & \textbf{9.4E-08} & - \\
 & 40\% & RMSE & \textbf{8.6E-02} & \textbf{1.5E-12} & \textbf{1.8E-07} & \textbf{6.2E-13} & \textbf{1.1E-10} & \textbf{3.0E-03} & \textbf{8.5E-12} & \textbf{2.7E-09} & \textbf{1.1E-02} & \textbf{1.3E-06} & \textbf{4.6E-07} & - \\
 &  & MAE & 2.2E-01 & \textbf{2.3E-12} & \textbf{4.0E-07} & \textbf{6.6E-13} & \textbf{2.9E-10} & \textbf{3.4E-03} & \textbf{7.8E-12} & \textbf{2.8E-09} & \textbf{5.8E-04} & \textbf{2.4E-06} & \textbf{4.2E-07} & - \\ \midrule
CN & 10\% & RMSE & \textbf{5.6E-02} & \textbf{2.2E-13} & \textbf{5.2E-11} & \textbf{2.2E-11} & \textbf{2.6E-12} & \textbf{7.3E-09} & \textbf{4.8E-09} & \textbf{4.9E-14} & \textbf{2.5E-09} & \textbf{4.0E-10} & \textbf{3.3E-09} & - \\
 &  & MAE & \textbf{2.3E-12} & \textbf{1.8E-16} & \textbf{2.1E-12} & \textbf{2.2E-14} & \textbf{2.3E-17} & \textbf{1.3E-11} & \textbf{8.9E-14} & \textbf{1.0E-15} & \textbf{1.3E-12} & \textbf{3.5E-10} & \textbf{1.9E-10} & - \\
 & 20\% & RMSE & \textbf{5.7E-03} & \textbf{2.1E-14} & \textbf{4.1E-12} & \textbf{1.3E-13} & \textbf{1.4E-14} & \textbf{3.2E-10} & \textbf{5.8E-11} & \textbf{2.1E-12} & \textbf{5.8E-11} & \textbf{7.3E-09} & \textbf{1.6E-10} & - \\
 &  & MAE & \textbf{5.2E-13} & \textbf{5.4E-16} & \textbf{2.2E-12} & \textbf{2.8E-15} & \textbf{2.3E-17} & \textbf{2.9E-12} & \textbf{9.3E-14} & \textbf{1.1E-13} & \textbf{9.3E-10} & \textbf{8.6E-08} & \textbf{1.0E-10} & - \\
 & 30\% & RMSE & \textbf{3.1E-05} & \textbf{1.2E-15} & \textbf{1.4E-12} & \textbf{2.7E-14} & \textbf{2.6E-14} & \textbf{1.7E-10} & \textbf{4.4E-12} & \textbf{7.2E-12} & \textbf{2.2E-09} & \textbf{2.3E-06} & \textbf{3.7E-12} & - \\
 &  & MAE & \textbf{8.5E-12} & \textbf{2.1E-17} & \textbf{5.9E-12} & \textbf{2.3E-16} & \textbf{2.6E-15} & \textbf{6.5E-12} & \textbf{3.4E-15} & \textbf{1.6E-13} & \textbf{1.1E-08} & \textbf{1.2E-05} & \textbf{2.2E-13} & - \\
 & 40\% & RMSE & \textbf{7.3E-09} & \textbf{6.6E-17} & \textbf{6.7E-13} & \textbf{1.1E-15} & \textbf{2.3E-13} & \textbf{2.3E-10} & \textbf{1.0E-13} & \textbf{7.5E-13} & \textbf{9.0E-10} & \textbf{7.9E-14} & \textbf{3.2E-12} & - \\
 &  & MAE & \textbf{1.4E-10} & \textbf{1.1E-18} & \textbf{1.5E-10} & \textbf{4.6E-16} & \textbf{2.0E-14} & \textbf{1.1E-10} & \textbf{2.0E-15} & \textbf{2.4E-14} & \textbf{4.0E-07} & \textbf{1.3E-13} & \textbf{5.6E-13} & - \\ \midrule
Los & 10\% & RMSE & \textbf{1.6E-07} & \textbf{2.3E-14} & \textbf{7.3E-13} & \textbf{3.7E-11} & \textbf{2.8E-13} & \textbf{1.4E-06} & \textbf{2.1E-09} & \textbf{3.1E-11} & \textbf{1.0E-05} & \textbf{3.2E-11} & \textbf{5.1E-09} & - \\
 &  & MAE & \textbf{1.3E-08} & \textbf{1.3E-13} & \textbf{9.5E-14} & \textbf{1.5E-12} & \textbf{4.2E-15} & \textbf{8.7E-07} & \textbf{3.0E-04} & \textbf{1.7E-13} & \textbf{1.6E-07} & \textbf{7.0E-13} & \textbf{4.8E-08} & - \\
 & 20\% & RMSE & \textbf{7.2E-09} & \textbf{2.2E-14} & \textbf{1.1E-12} & \textbf{2.7E-12} & \textbf{2.1E-15} & \textbf{9.9E-07} & \textbf{2.9E-09} & \textbf{4.3E-11} & \textbf{8.7E-06} & \textbf{8.9E-12} & \textbf{9.3E-13} & - \\
 &  & MAE & \textbf{1.9E-10} & \textbf{7.5E-15} & \textbf{3.4E-13} & \textbf{1.4E-13} & \textbf{1.9E-17} & \textbf{7.8E-07} & \textbf{3.5E-04} & \textbf{7.7E-12} & \textbf{8.9E-06} & \textbf{1.2E-12} & \textbf{1.0E-10} & - \\
 & 30\% & RMSE & \textbf{7.9E-10} & \textbf{1.0E-15} & \textbf{9.9E-13} & \textbf{3.3E-10} & \textbf{1.0E-14} & \textbf{1.4E-07} & \textbf{2.8E-09} & \textbf{9.9E-15} & \textbf{4.6E-05} & \textbf{2.0E-12} & \textbf{3.7E-10} & - \\
 &  & MAE & \textbf{1.8E-11} & \textbf{1.5E-15} & \textbf{8.2E-14} & \textbf{7.5E-12} & \textbf{1.0E-15} & \textbf{3.6E-07} & \textbf{3.5E-04} & \textbf{7.7E-13} & \textbf{5.1E-05} & \textbf{2.0E-12} & \textbf{2.7E-07} & - \\
 & 40\% & RMSE & \textbf{7.7E-05} & \textbf{4.2E-17} & \textbf{1.8E-13} & \textbf{1.7E-10} & \textbf{3.8E-15} & \textbf{3.3E-08} & \textbf{3.1E-09} & \textbf{1.3E-12} & \textbf{1.9E-05} & \textbf{1.8E-10} & \textbf{4.9E-11} & - \\
 &  & MAE & \textbf{3.9E-05} & \textbf{3.8E-16} & \textbf{2.9E-12} & \textbf{6.7E-12} & \textbf{1.8E-16} & \textbf{2.5E-08} & \textbf{3.8E-04} & \textbf{2.1E-13} & \textbf{1.2E-05} & \textbf{9.5E-12} & \textbf{7.4E-08} & - \\ \midrule
LuohuTaxi & 10\% & RMSE & \textbf{2.0E-09} & \textbf{1.2E-11} & \textbf{1.1E-08} & \textbf{1.7E-08} & \textbf{1.8E-11} & \textbf{8.1E-06} & \textbf{4.7E-04} & \textbf{2.2E-05} & \textbf{4.8E-08} & \textbf{3.4E-09} & \textbf{2.2E-05} & - \\
 &  & MAE & \textbf{9.7E-10} & \textbf{9.5E-14} & \textbf{2.3E-09} & \textbf{1.6E-09} & \textbf{3.5E-12} & \textbf{1.5E-06} & \textbf{8.4E-05} & \textbf{6.7E-07} & \textbf{2.0E-06} & \textbf{2.3E-10} & \textbf{2.0E-05} & - \\
 & 20\% & RMSE & \textbf{4.0E-10} & \textbf{1.1E-14} & \textbf{3.7E-11} & \textbf{4.3E-11} & \textbf{6.6E-13} & \textbf{2.1E-08} & \textbf{4.5E-04} & \textbf{1.5E-05} & \textbf{1.0E-10} & \textbf{5.8E-11} & \textbf{1.1E-11} & - \\
 &  & MAE & \textbf{6.5E-11} & \textbf{3.7E-16} & \textbf{4.5E-10} & \textbf{3.0E-10} & \textbf{3.8E-14} & \textbf{7.6E-08} & \textbf{7.7E-05} & \textbf{3.7E-07} & \textbf{6.4E-08} & \textbf{1.1E-12} & \textbf{4.4E-09} & - \\
 & 30\% & RMSE & \textbf{2.3E-10} & \textbf{1.3E-14} & \textbf{2.0E-09} & \textbf{2.8E-14} & \textbf{3.8E-14} & \textbf{1.8E-08} & \textbf{4.6E-04} & \textbf{3.0E-05} & \textbf{1.0E-11} & \textbf{1.8E-10} & \textbf{2.7E-08} & - \\
 &  & MAE & \textbf{1.4E-09} & \textbf{2.7E-16} & \textbf{1.2E-10} & \textbf{2.8E-12} & \textbf{9.9E-16} & \textbf{1.1E-08} & \textbf{7.8E-05} & \textbf{1.3E-06} & \textbf{1.8E-09} & \textbf{1.6E-12} & \textbf{1.6E-08} & - \\
 & 40\% & RMSE & \textbf{1.4E-10} & \textbf{2.7E-16} & \textbf{2.0E-08} & \textbf{1.1E-14} & \textbf{1.3E-12} & \textbf{4.0E-08} & \textbf{5.1E-04} & \textbf{1.6E-05} & \textbf{4.5E-11} & \textbf{5.7E-12} & \textbf{5.8E-10} & - \\
 &  & MAE & \textbf{9.7E-10} & \textbf{7.1E-19} & \textbf{1.0E-09} & \textbf{3.1E-13} & \textbf{9.8E-14} & \textbf{1.1E-08} & \textbf{8.5E-05} & \textbf{1.0E-06} & \textbf{3.7E-10} & \textbf{1.1E-13} & \textbf{2.6E-08} & - \\
\bottomrule
\end{tabular}
    }
    \vspace{-0.15in}
    \caption{
    T-test P-value (bolding for significant with p=0.05) of the imputation performance between GSLI and existing methods  with various missing rates}
    \vspace{-0.2in}
    \label{tab:t-test-exp}
\end{table*}

\begin{table*}[h]
\centering
     \renewcommand\tabcolsep{1pt}
     \renewcommand{\arraystretch}{1}
     \resizebox{\linewidth}{!}{
     \renewcommand\tabcolsep{4pt}
\begin{tabular}{@{}lll|lllllllllllll@{}}
\toprule
Dataset & Missing rate & Metric & CSDI & TimesNet & SAITS & GPT4TS & LRTC-TNN & GRIN & STD-GAE & DAMR & PriSTI & PoGeVon & ImputeFormer & GSLI \\ \midrule
DutchWind & 10\% & RMSE & 11.67\% & 19.52\% & 13.38\% & 26.71\% & 29.99\% & 6.20\% & 13.30\% & 32.69\% & 15.03\% & 16.86\% & 22.08\% & - \\
 &  & MAE & 7.94\% & 35.56\% & 22.47\% & 44.02\% & 30.06\% & 10.59\% & 18.45\% & 48.51\% & 5.30\% & 28.52\% & 31.51\% & - \\
 & 20\% & RMSE & 13.94\% & 32.05\% & 12.59\% & 36.97\% & 31.05\% & 4.47\% & 14.01\% & 31.02\% & 13.74\% & 10.69\% & 22.23\% & - \\
 &  & MAE & 11.53\% & 50.18\% & 21.73\% & 54.51\% & 32.49\% & 8.82\% & 20.01\% & 47.25\% & 6.08\% & 16.68\% & 31.94\% & - \\
 & 30\% & RMSE & 13.71\% & 39.26\% & 12.58\% & 42.26\% & 33.89\% & 3.12\% & 15.83\% & 29.15\% & 14.11\% & 12.33\% & 20.96\% & - \\
 &  & MAE & 13.72\% & 56.69\% & 21.78\% & 59.01\% & 36.14\% & 6.94\% & 23.45\% & 45.23\% & 7.84\% & 20.82\% & 30.46\% & - \\
 & 40\% & RMSE & 14.81\% & 43.05\% & 11.12\% & 45.11\% & 36.37\% & 1.54\% & 20.50\% & 27.76\% & 16.51\% & 21.18\% & 20.58\% & - \\
 &  & MAE & 17.62\% & 59.48\% & 20.16\% & 60.96\% & 40.26\% & 5.11\% & 30.52\% & 43.02\% & 12.90\% & 35.63\% & 28.97\% & - \\ \midrule
BeijingMEO & 10\% & RMSE & 14.49\% & 16.22\% & 17.94\% & 24.38\% & 35.65\% & 7.69\% & 17.88\% & 44.89\% & 12.80\% & 25.30\% & 22.80\% & - \\
 &  & MAE & 2.17\% & 33.62\% & 29.85\% & 43.22\% & 36.83\% & 15.80\% & 30.17\% & 60.15\% & 4.52\% & 42.40\% & 28.12\% & - \\
 & 20\% & RMSE & 14.83\% & 22.93\% & 17.59\% & 35.84\% & 38.15\% & 7.17\% & 17.55\% & 44.22\% & 13.83\% & 24.76\% & 22.86\% & - \\
 &  & MAE & 3.16\% & 42.39\% & 30.64\% & 56.50\% & 38.75\% & 15.11\% & 29.27\% & 59.35\% & 8.02\% & 39.36\% & 28.68\% & - \\
 & 30\% & RMSE & 15.38\% & 30.83\% & 16.71\% & 42.42\% & 40.20\% & 6.79\% & 17.03\% & 41.82\% & 17.77\% & 22.45\% & 23.33\% & - \\
 &  & MAE & 4.57\% & 50.67\% & 29.18\% & 62.00\% & 40.54\% & 14.54\% & 27.98\% & 56.90\% & 17.24\% & 34.83\% & 29.42\% & - \\
 & 40\% & RMSE & 16.37\% & 37.15\% & 16.26\% & 46.28\% & 42.32\% & 6.58\% & 16.68\% & 40.85\% & 24.66\% & 24.61\% & 21.78\% & - \\
 &  & MAE & 6.80\% & 56.36\% & 29.16\% & 64.75\% & 42.54\% & 14.60\% & 26.83\% & 55.94\% & 25.46\% & 38.39\% & 27.87\% & - \\ \midrule
LondonAQ & 10\% & RMSE & 8.83\% & 32.94\% & 27.55\% & 43.45\% & 44.49\% & 12.68\% & 54.46\% & 62.28\% & 13.40\% & 27.55\% & 32.34\% & - \\
 &  & MAE & 5.11\% & 34.52\% & 30.52\% & 46.17\% & 44.19\% & 12.49\% & 63.31\% & 64.94\% & 9.86\% & 25.42\% & 34.00\% & - \\
 & 20\% & RMSE & 4.89\% & 43.35\% & 23.42\% & 51.82\% & 42.66\% & 8.14\% & 49.88\% & 58.98\% & 23.91\% & 22.74\% & 26.01\% & - \\
 &  & MAE & 1.16\% & 47.59\% & 27.48\% & 56.26\% & 43.80\% & 7.96\% & 60.47\% & 62.38\% & 8.90\% & 21.68\% & 29.33\% & - \\
 & 30\% & RMSE & 5.73\% & 51.36\% & 22.48\% & 56.75\% & 45.20\% & 8.64\% & 51.82\% & 59.28\% & 36.20\% & 22.89\% & 26.50\% & - \\
 &  & MAE & 3.73\% & 57.34\% & 27.73\% & 62.28\% & 47.89\% & 9.99\% & 62.44\% & 63.34\% & 18.08\% & 23.42\% & 30.48\% & - \\
 & 40\% & RMSE & 10.89\% & 55.21\% & 21.92\% & 58.76\% & 47.87\% & 7.06\% & 47.01\% & 57.19\% & 46.35\% & 21.82\% & 22.82\% & - \\
 &  & MAE & 3.29\% & 60.48\% & 26.07\% & 63.58\% & 49.29\% & 8.07\% & 58.54\% & 60.62\% & 26.76\% & 21.92\% & 26.46\% & - \\ \midrule
CN & 10\% & RMSE & 46.28\% & 62.08\% & 46.60\% & 48.29\% & 54.82\% & 37.08\% & 31.50\% & 71.16\% & 34.53\% & 60.03\% & 34.80\% & - \\
 &  & MAE & 33.84\% & 73.76\% & 57.84\% & 61.84\% & 66.36\% & 50.08\% & 41.40\% & 80.37\% & 32.72\% & 69.52\% & 41.07\% & - \\
 & 20\% & RMSE & 38.78\% & 61.75\% & 44.43\% & 47.56\% & 55.14\% & 35.88\% & 31.19\% & 69.77\% & 36.77\% & 57.79\% & 33.31\% & - \\
 &  & MAE & 33.81\% & 73.46\% & 55.48\% & 61.30\% & 66.75\% & 48.69\% & 40.42\% & 79.31\% & 34.79\% & 67.34\% & 38.64\% & - \\
 & 30\% & RMSE & 36.15\% & 61.46\% & 42.43\% & 48.11\% & 56.16\% & 34.91\% & 30.23\% & 68.83\% & 40.76\% & 55.30\% & 32.49\% & - \\
 &  & MAE & 34.33\% & 73.06\% & 53.33\% & 61.96\% & 67.41\% & 47.35\% & 39.20\% & 77.92\% & 38.64\% & 65.21\% & 37.78\% & - \\
 & 40\% & RMSE & 35.60\% & 60.91\% & 39.98\% & 49.53\% & 57.61\% & 33.69\% & 29.64\% & 66.92\% & 41.56\% & 48.55\% & 32.52\% & - \\
 &  & MAE & 35.36\% & 72.39\% & 50.56\% & 63.26\% & 68.08\% & 45.64\% & 38.07\% & 76.07\% & 42.51\% & 56.87\% & 36.43\% & - \\ \midrule
Los & 10\% & RMSE & 15.50\% & 50.45\% & 50.78\% & 33.68\% & 47.47\% & 10.85\% & 72.16\% & 48.72\% & 10.15\% & 27.95\% & 40.81\% & - \\
 &  & MAE & 10.20\% & 53.01\% & 45.56\% & 38.04\% & 51.98\% & 15.19\% & 70.56\% & 56.14\% & 14.61\% & 23.86\% & 29.90\% & - \\
 & 20\% & RMSE & 17.94\% & 51.20\% & 49.18\% & 39.56\% & 50.88\% & 10.72\% & 70.98\% & 46.92\% & 12.73\% & 27.42\% & 38.17\% & - \\
 &  & MAE & 11.56\% & 55.19\% & 44.77\% & 44.99\% & 56.01\% & 15.60\% & 69.67\% & 54.76\% & 18.34\% & 24.11\% & 26.60\% & - \\
 & 30\% & RMSE & 20.59\% & 53.24\% & 48.28\% & 48.16\% & 54.45\% & 11.70\% & 70.19\% & 45.99\% & 20.17\% & 27.84\% & 38.70\% & - \\
 &  & MAE & 13.27\% & 58.19\% & 43.93\% & 52.90\% & 59.63\% & 16.61\% & 68.99\% & 54.44\% & 25.42\% & 24.80\% & 27.07\% & - \\
 & 40\% & RMSE & 26.19\% & 54.73\% & 47.35\% & 54.91\% & 55.95\% & 11.66\% & 68.89\% & 44.09\% & 28.10\% & 26.86\% & 36.32\% & - \\
 &  & MAE & 17.51\% & 60.69\% & 43.38\% & 58.99\% & 61.92\% & 16.83\% & 68.00\% & 52.37\% & 35.08\% & 24.85\% & 25.10\% & - \\ \midrule
LuohuTaxi & 10\% & RMSE & 12.19\% & 20.24\% & 9.22\% & 28.82\% & 39.67\% & 6.01\% & 47.59\% & 38.52\% & 21.59\% & 17.40\% & 10.00\% & - \\
 &  & MAE & 11.70\% & 28.78\% & 10.46\% & 35.65\% & 39.79\% & 8.28\% & 54.91\% & 47.17\% & 24.70\% & 18.80\% & 10.73\% & - \\
 & 20\% & RMSE & 11.52\% & 23.29\% & 8.89\% & 40.17\% & 42.32\% & 5.70\% & 46.94\% & 38.29\% & 21.17\% & 16.85\% & 8.50\% & - \\
 &  & MAE & 11.38\% & 32.70\% & 10.45\% & 44.50\% & 42.91\% & 8.09\% & 54.33\% & 46.96\% & 22.98\% & 18.41\% & 9.33\% & - \\
 & 30\% & RMSE & 11.32\% & 27.38\% & 8.09\% & 45.80\% & 45.35\% & 5.52\% & 46.35\% & 38.77\% & 21.18\% & 16.73\% & 8.22\% & - \\
 &  & MAE & 11.38\% & 37.09\% & 9.46\% & 49.74\% & 46.59\% & 7.91\% & 53.79\% & 47.45\% & 23.54\% & 18.21\% & 9.13\% & - \\
 & 40\% & RMSE & 10.58\% & 31.76\% & 8.00\% & 48.81\% & 45.67\% & 5.31\% & 45.75\% & 38.04\% & 21.02\% & 17.00\% & 7.16\% & - \\
 &  & MAE & 11.20\% & 41.23\% & 9.62\% & 52.84\% & 47.64\% & 7.75\% & 53.16\% & 46.80\% & 23.20\% & 18.26\% & 7.64\% & -\\
\bottomrule
\end{tabular}
    }
    \vspace{-0.15in}
    \caption{
        Improvement percentage of the imputation performance between existing methods and GSLI with various missing rates
  }
    \vspace{-0.2in}
    \label{tab:improve-percentage-exp}
\end{table*}

\end{document}